\documentclass[10pt,twocolumn,letterpaper]{article}

\usepackage{iccv}              
\usepackage{multirow}

\definecolor{iccvblue}{rgb}{0.21,0.49,0.74}
\usepackage[pagebackref,breaklinks,colorlinks,allcolors=iccvblue]{hyperref}

\def\paperID{10002} 
\def\confName{ICCV}
\def\confYear{2025}

\title{ARIG: Autoregressive Interactive Head Generation for Real-time Conversations}

\author{Ying Guo\textsuperscript{*}, Xi Liu\thanks{Equal Contribution}, Cheng Zhen\thanks{Corresponding Author}, Pengfei Yan, Xiaoming Wei  \vspace{2mm}\\
Vision AI Department, Meituan\\
\small \url{https://jinyugy21.github.io/ARIG/}
}

\begin{document}
\maketitle
\begin{abstract}
Face-to-face communication, as a common human activity, motivates the research on interactive head generation. A virtual agent can generate motion responses with both listening and speaking capabilities based on the audio or motion signals of the other user and itself. However, previous clip-wise generation paradigm or explicit listener/speaker generator-switching methods have limitations in future signal acquisition, contextual behavioral understanding, and switching smoothness, making it challenging to be real-time and realistic.
In this paper, we propose an autoregressive (AR) based frame-wise framework called ARIG to realize the real-time generation with better interaction realism.
To achieve real-time generation, we model motion prediction as a non-vector-quantized AR process. Unlike discrete codebook-index prediction, we represent motion distribution using diffusion procedure, achieving more accurate predictions in continuous space.
To improve interaction realism, we emphasize interactive behavior understanding (IBU) and detailed conversational state understanding (CSU). In IBU, based on dual-track dual-modal signals, we summarize short-range behaviors through bidirectional-integrated learning and perform contextual understanding over long ranges. In CSU, we use voice activity signals and context features of IBU to understand the various states (interruption, feedback, pause, etc.) that exist in actual conversations. These serve as conditions for the final progressive motion prediction.
Extensive experiments have verified the effectiveness of our model.
\end{abstract}    
\section{Introduction}
\label{sec:intro}

Face-to-face communication is a common human activity around us \cite{dialoguenerf}. Research on head generation makes it possible for virtual agents to become communication participants.
Early works focus on single-role generation (talking/listening). 
In talking head generation, the input signal comes only from the speaker himself and does not receive signals from the other participant for interaction.
The listening head generation receives signals from both participants to provide appropriate listener responses, but it only offers non-verbal motion feedback and cannot generate speaker responses. 
In real conversations, each participant's response should integrate the behaviors of both parties and possess the ability to both speak and listen.
Thus, single-role generation cannot fully simulate behaviors in conversations.

\begin{figure}[t]
   \includegraphics[width=\linewidth]{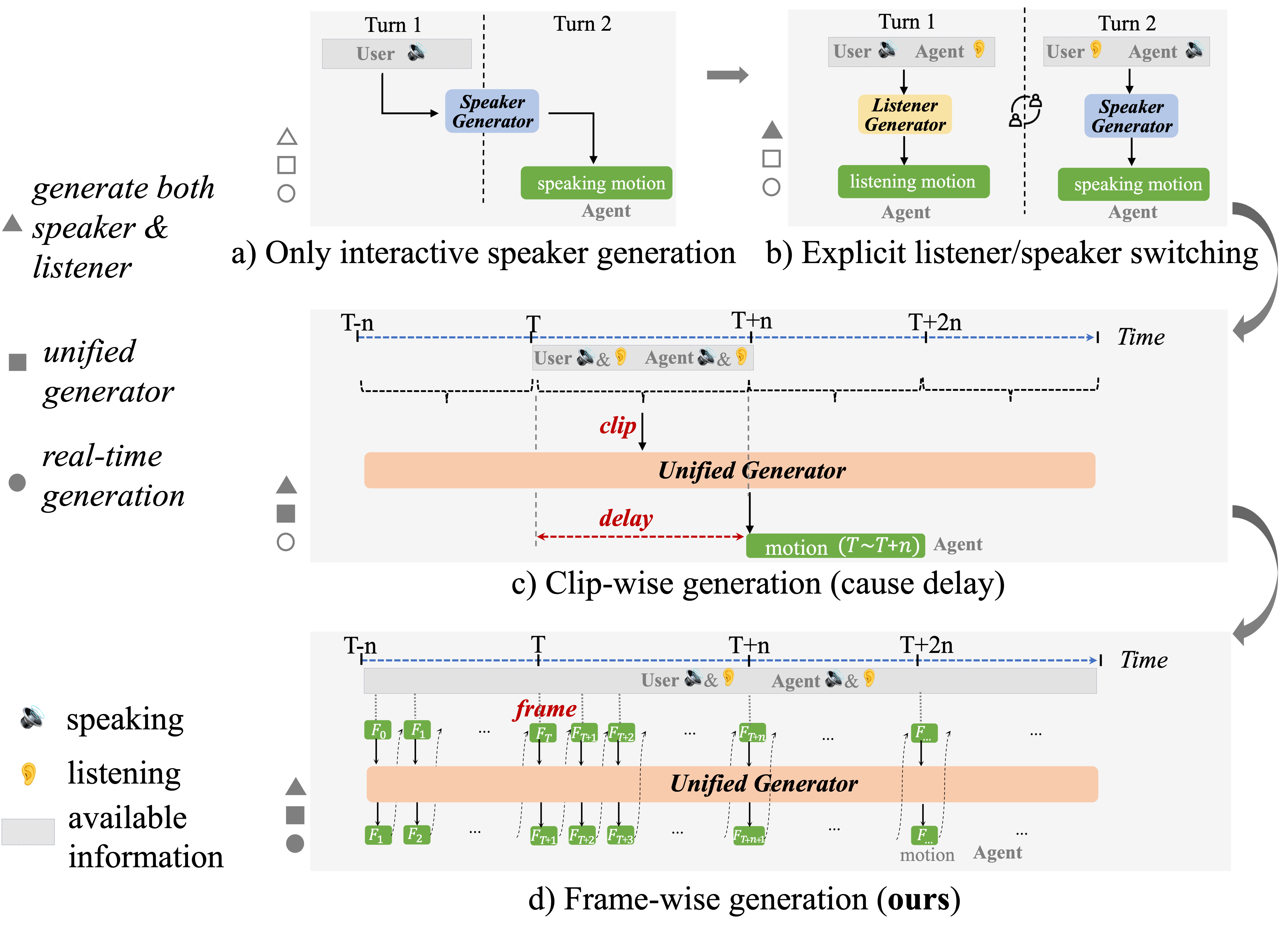}
   \vspace{-0.5cm}
   \caption{Development of interactive head motion generation. a) Only speaker generation based on last speaking behaviors. b) Different explicit generators for listener/speaker. c) Implicit clip-by-clip  generator based on a unified model. d) Implicit frame-by-frame generator based on a unified non-quantized AR model.}
   \vspace{-0.5cm}
   \label{fig:figure_1}
\end{figure}

Recently, some works \cite{dim,infp,mrecgen,vicox,wang2023agentavatar,park2024let} have studied the interactive head generation (IHG), as shown in \cref{fig:figure_1}. 
Suppose the conversation between a user and a virtual agent, IHG generates interactive head responses for the agent by utilizing available information from both parties.
Initially, early works (\cref{fig:figure_1}a) such as MRecGen \cite{mrecgen} and \cite{park2024let} model the conversation as a multi-turn mode and only generate the agent's speaker response based on the user's speaking behavior in the last turn. 
Then, methods (\cref{fig:figure_1}b) in \cite{vicox}, DIM \cite{dim} and AgentAvatar \cite{wang2023agentavatar} can generate interactive responses of both speakers and listeners.
However, they have to assign explicit roles (speaker/listener) for the agent and switch between two respective generators.
Conversations involve more complex states
beyond speaker/listener alternation (\eg interruption, pause, giving verbal feedback), requiring immediate and smooth role switching, and sometimes even both parties play the same role, which is difficult to achieve with explicit role switching.
To solve this, in \cref{fig:figure_1}c, INFP \cite{infp} uses a unified generator to switch roles implicitly and dynamically.
Although its inference speed is ideal, it is a clip-wise paradigm that inputs two fixed-length clips ({\small$T\!\!\sim \!\!T\!+n$}) of audio and output motions within {\small$T\!\!\sim \!\!T\!+n$}. 
At time $T$, since the signal about the future $T\!+\!n$ cannot be acquired, the model must wait until the audio is acquired at {\small$T\!+\!n$} and then forms a clip to generate the motion at $T$,
which will cause a delay of at least one clip, making \textit{real-time generation} difficult.

Besides, for \textit{interaction realism}, the clip-wise paradigm limits the visible range to short clips. Without long context, it may misinterpret the logic of statements and the person's intent, resulting in unreasonable motions.
Moreover, its input in motion guider contains only audio without visual information, which may not be sufficient to fully represent the behavior of a conversation participant, who may interact in many non-audio ways such as frowning and nodding.

To solve these problems, we propose a novel framework called \textbf{ARIG} based on \textbf{A}uto\textbf{R}egression to redefine the \textbf{I}nteraction \textbf{G}eneration paradigm which is \textit{real-time} and has better \textit{interaction realism}.
\textit{For real-time generation}, we design a frame-wise generation framework (\cref{fig:figure_1}d) which generates the $T$-$th$ frame promptly in an autoregressive (AR) manner based on previous $T$-$1$ frames, 
which can respond in time,
without future clip being available.
We implement it with a continuous non-quantized AR, enjoying the advantages of sequential processing and more accurate motion representation.
\textit{For better interaction realism},
we emphasize the impact of contextual semantics by combining long- and short-range features in AR sequences
and integrate dual-track (user and agent) dual-modal (audio and visual motion) signals to understand interactive behaviors.
Furthermore, we analyze complex conversion states (interrupted, feedback, wait, overlap, pause, etc.), noting that even with the same audio, the motions differ in different states.
For example, for ``wow it's amazing to...", in a normal speaking state, one might widen eyes to express exclamation, while in a state of giving feedback to the other party, nodding may be performed as approval. Thus, we learn the state features as a guide for motion generation.

Specifically, 
\textbf{to achieve real-time frame-wise generation with higher fidelity}, we model the framework as non-quantized autoregression (AR) for continuous-valued predictions. 
Unlike conventional AR motion modeling that predicts discrete and finite codebook indices, we use a diffusion procedure to model the probability distribution of motion, enabling the prediction of continuous values to reduce deviation of discretization and better reproduce micro-expressions.
\textbf{To better understand the contextual interaction}, we design an interactive behavior understanding module (IBU).
Considering the learning difficulties and time costs of excessive long-range information,
we first effectively summarizes dual-modal dual-track interaction by bidirectional-integrated learning within a short chunk, and perform contextual understanding over a long range.
\textbf{To comprehensively learn complex conversation states}, we design a conversation states understanding module (CSU), which combines the voice activity signal and the contextual feature of IBU module to learn current conversation states.
Finally, we conduct a progressive motion prediction (PMP) conditioned on context, conversation states, and audio, combining past motions to generate final motions.

In summary, this paper has the following contributions:
\begin{itemize}
\setlength{\itemsep}{0pt}
\item We propose a novel frame-wise generation framework based on continuous AR with diffusion probability modeling, which formulates the real-time paradigm of interactive head generation with higher motion fidelity.
\item 
We design an IBU module to learn the contextual interaction, which
effectively summarize dual-track dual-modal signals of short range via bidirectional-integrated learning, and capture contextual information over a long range.
\item We analyze the complex states in real-world conversations in more detail, and design a CSU module to learn the state feature to guide the motion generation.
\item Extensive experiments demonstrate that our method has significant improvements in real-time realistic interactive head generation as well as single-role generation, and we provide the full effects in the Supplementary Video.
\end{itemize}

\section{Related Work}
\label{sec:related}
\vspace{-0.1cm}
\subsection{Single-role Head Generation}
\vspace{-0.1cm}
\paragraph{Talking head generation (THG):} Audio-driven THG is to generate the speaking video with verbal-motions given driving audio and a reference face image.
Some methods \cite{tian2024emo,jiang2024loopy,zheng2024memo} realize head animations by fine-tuning the pre-trained diffusion models.
Hallo\cite{xu2024hallo} uses a hierarchical audio-driven visual synthesis module to enhance precision and EchoMimic\cite{chen2024echomimic} combines both audios and facial landmarks.
In general, THG receives audios solely from the speaker's side, and focuses on audio-lip sync and expression-audio alignment for talking generation.
\vspace{-0.5cm}
\paragraph{Listening head generation (LHG):}LHG is to generate listener responses synchronized with the speaker based on the speaker's audio and motions, as well as the listener's own past motions.
Some works, such as RLHG \cite{vico}, PCH \cite{pchg}, L2L \cite{l2l}, ELP \cite{ELP}, and MFR-Net \cite{mfr} generate head motions based on 3D face coefficients, and CustomListener \cite{liu2024customlistener} further control motions through a text-prior guidance.
LHG provides non-verbal responses synchronized with the speaker, but cannot produce verbal motions.

\begin{figure*}[t]
   \includegraphics[width=\linewidth]{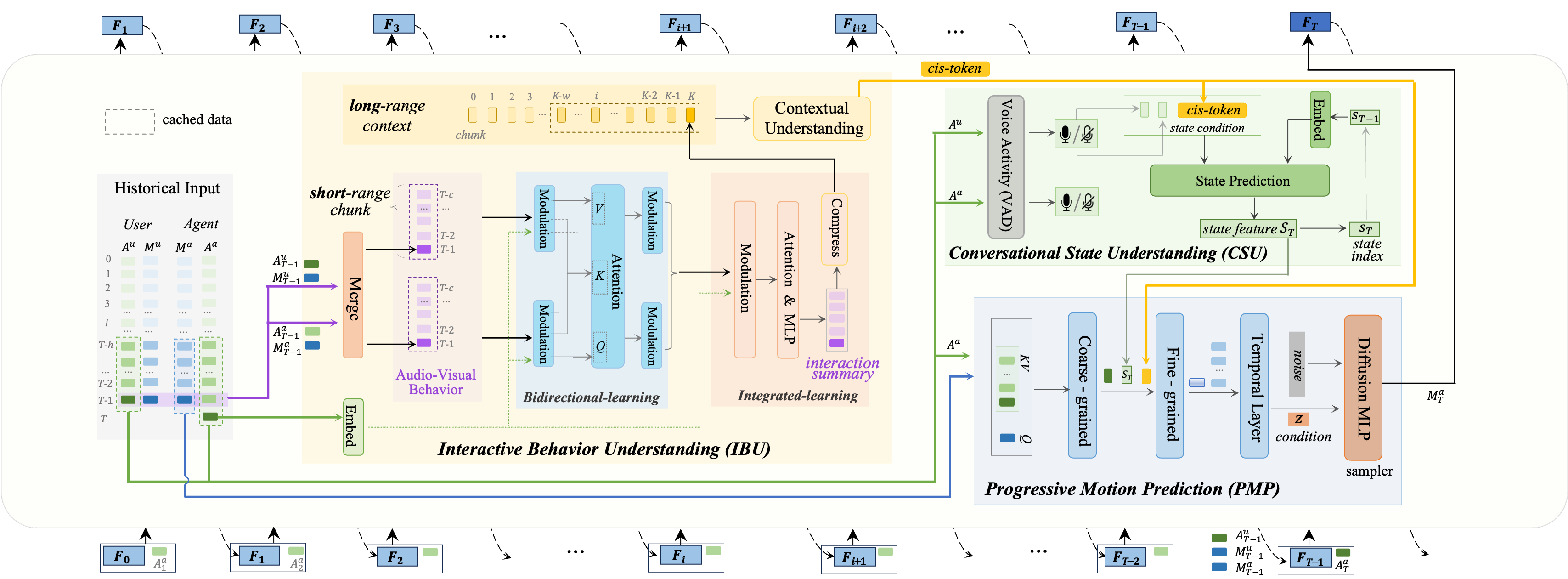}
   \caption{Overall framework of ARIG. Given the previous frame's audio and motion $A_{T\!-\!1}^u$, $M_{T\!-\!1}^a$, $M_{T\!-\!1}^u$ along with the current audio $A_{T}^a$, IBU first effectively uses dual-track dual-modal signals to perform bidirectional-integrated learning over short ranges and contextual understanding over long ranges, obtaining a contextual interaction summary (cis-token). Then, CSU combines voice activity and cis-token to predict state features. Finally, PMP progressively predicts motions and uses the DiffusionMLP to sample the final motion.}
   \vspace{-0.5cm}
   \label{fig:main}
\end{figure*}

\vspace{-0.2cm}
\subsection{Interactive Head Generation (IHG)}
\vspace{-0.2cm}
In IHG, some works \cite{dialoguenerf,huang2024interact,ng2024audio,sun2024beyond} have studied person-specific generation.
For generalized arbitrary identities, MRecGen \cite{mrecgen} and \cite{park2024let} process the conversation as a multi-turn pattern and generate the agent's speaker response based on the user's speaking behavior in the last turn, without the generation of synchronous listener behavior.
Then, some works \cite{dim,vicox,wang2023agentavatar} achieve both the role of speaker and listener.
Zhou \etal \cite{vicox} utilizes two independent generators and switch generators through a role switcher network. 
DIM \cite{dim} uses the pretraining strategy to encode a unified representation of both roles and adapts to the single role by downstream masking and fine-tuning. 
AgentAvatar \cite{wang2023agentavatar} designs prompts to produce detailed motion descriptions through LLMs, which are then processed into motion token sequences, and further generate the final animations via rendering. 
Although they can generate both roles, they need explicit role pre-assignment to match respective generators, which is not smooth enough when switching immediately and cannot adapt to situations where both parties have the same role.
Furthermore, INFP \cite{infp} uses a unified generator to learn the dynamic state, without explicit role assignments. 
However, its clip-wise paradigm makes it difficult to achieve real-time generation, and learning based on a short clip and only audio signals limits its understanding of long-range contextual information and the overall behavior of both parties.
For our ARIG, we propose a real-time frame-wise paradigm, and introduce long range understanding based on dual-modal inputs to achieve better interaction.

\vspace{-0.2cm}
\subsection{Autoregressive Motion Generation}
\vspace{-0.2cm}
Autoregressive (AR) models have the nature of ``predicting next tokens based on known ones” \cite{mar}.
Influenced by the discrete vocabulary of language models, common AR Motion Generation pre-trains a discrete-valued tokenizer, which involves a finite codebook obtained by vector quantization (VQ) \cite{sun2024beyond,zhou2023unified}, and then learn the correlation between the codebook indices.
However, discrete VQ is not as accurate as continuous values in representing detailed motions and is sensitive to gradient estimation \cite{van2017neural,ramesh2021zero,kolesnikov2022uvim}.
Recently, some works \cite{mar,deng2024nova}
study non-VQ AR in continuous-valued space. 
Inspired by this, we model the per-motion probability using a diffusion procedure instead of discrete categorical distribution.
which enjoys both the sequence processing ability of AR and the accuracy of continuous values.
Note that although \cite{ng2024audio} integrates diffusion in AR, it merely concatenates VQ and diffusion and does not uniformly model motion as a diffusion distribution, and its diffusion input remains in clip form instead of frame by frame.

\section{Method}
\subsection{Overview}
In our ARIG framework, we integrate the dual-modal (i.e. audio and visual motion) signals from dual-track (i.e. user and agent) to generate the agent's motions frame by frame.
We use motion coefficients extracted by \cite{liveportrait} to represent head motions, which include expression, pose, and scale information, and use optical-flow based generation \cite{liveportrait} to obtain videos.
Let $M_t^a$ and $M_t^u$ represent the agent's and user's motions at time $t$, and $A_t^a$ and $A_t^u$ represent their audios.
When generating the $T$-$th$ frame, the audio and motions of the user and agent in previous $T$-$1$ frames, as well as the agent's audio at current $T$ are available. They are used as input to generate the agent's motion $M_T^a$. This process can be formulated as:
\begin{equation}
M_T^a=\operatorname{ARIG}\left(A_{0 \sim T}^a, A_{0 \sim T\!-\!1}^u, M_{0 \sim T\!-\!1}^a, M_{0 \sim T\!-\!1}^u\right)
\end{equation}
In our AR process, each update involves data at $T\!-\!1$ and $T$, while the historical data has been selectively utilized, transformed into model features, and cached. Therefore, the actual process can be reformulated as:
\begin{equation}
M_T^a=\operatorname{ARIG}\left(A_{T}^a, A_{T\!-\!1}^u, M_{T\!-\!1}^a, M_{T\!-\!1}^u\right)
\end{equation}

The overview of our framework is shown in \cref{fig:main}.
First, in IBU, we merge audio and motion and perform contextual understanding via the long-short combination. We first divide time into short-range chunks of window size $c$, and learn the mutual interaction behavior within the chunk by the bidirectional-integrated structure and obtain a single summary. Then we understand contextual behaviors in a long range. The context is composed of chunk's compressed interaction summary, which provides a more concise and clearer perspective, and effectively increases the information capacity of the long range.
In CSU, we first obtain the current voice activity signals of both parties through VAD, and then learn the current state representations from both voice activity signals and semantics of IBU.
Then in PMP, we first predict a coarse motion based on the audio and then refine it with state, context, and audio for fine-grained adjustment. We use temporal layers to ensure inter-frame continuity, and finally predict motions in continuous space by modeling diffusion distributions in DiffusionMLP.

\subsection{Interactive Behavior Understanding Module}
In this IBU module, we focus mainly on how to effectively utilize dual-modal (audio and visual-motion) information from dual-track (user and agent) participants to understand interactive behaviors.
In general, we adopt a combination of long and short ranges. Understanding long-range context is crucial for better capturing contextual semantic information; otherwise, the agent might fail to grasp the tone of a character, misunderstand the logical relationships between sentences, or misinterpret a person's intent based solely on short audio snippets.
However, excessive information over long ranges can lead to computational cost and memory usage. Besides, learning independent, scattered information frame by frame also poses challenges for effective understanding.
Therefore, we first summarize the interaction information within a short period of time, and then conduct long-range learning based on the short-range summary. This can effectively condense information to improve efficiency, and using short chunks as units can help the content understanding from a more concise and clear perspective.

For dual-modal input, we believe that only by combining audio and visual motions can we accurately represent a person's behavior. For example, silence with a frown or an angry tone without obvious expression are both negative behaviors, and focusing on a single modality may lead to misunderstandings.
So we first use an MLP-based merge block to combine the information of the two modalities to form a comprehensive audio-visual behavior for each frame. 
Specifically at the generation of the $T$-$th$ frame, we can obtain audio-visual behaviors $I_{T\!-\!1}^a$ and $I_{T\!-\!1}^u$ based on the updated $A_{T\!-\!1}^u, M_{T\!-\!1}^a, M_{T\!-\!1}^u$ and the historical $A_{T\!-\!1}^a$.

\vspace{-0.5cm}
\paragraph{Short-range understanding}
We first divide the time into chunks of window size $c$, i.e., each chunk corresponds to $c$ frames. For $t$-$th$ frame, the index of the corresponding chunk is $i=\lfloor t / c\rfloor$, and the frame range corresponding to each chunk is $[c i, c(i+1))$, where $\lfloor\cdot\rfloor$ is the floor operation.
We use 2 chunk caches for 2 parties to store the frame-level audio-visual behaviors of the latest chunk and dynamically update them with a sliding window, which are the behavior sets of the user and the agent for subsequent interaction understanding via bidirectional-integrated learning.

In bidirectional learning, we first consider the user and the agent as independent individuals, understand their own information based on the audio updated at time $T$, and exchange information between each other through the attention mechanism. This is inspired by the way MMDiT \cite{mmdit} handles text and image modalities.
In each independent channel, we use the modulation mechanism composed of adaptive LayerNorm and Linear layer to learn their own behavior, and use the shared attention mechanism to supplement respective understanding with the other party's information.
Then, in integrated learning, we concatenate the dual-track output and integrate the two parties' behavior understanding to summarize the interaction in the current chunk. We also use a modulation mechanism to inject updated audio, so as to enhance the information fusion for interaction understanding conditioned on audio. 
We use the parallel Attention-MLP \cite{dehghani2023scaling} to improve efficiency. 
In this way, we effectively achieve the cooperation of the five types of signals, including dual-track dual-modal information in the cache and the updated audio, so that they can fully understand the information between each other and obtain the interaction summary. 
Then we use a linear layer to compress it and put it into the long-range context cache.

\vspace{-0.5cm}
\paragraph{Long-range understanding}
We store the summary of each chunk in a context cache with a capacity $w$.
Contextual understanding is achieved through a decoder-only structure with the causal mask, obtaining the final contextual interaction summary (cis-token).
Note that when the frames in the chunk cache do not completely belong to the current chunk $i$, that is, when the frames of the current chunk $i$ are insufficient, we will refresh the summary of the chunk $i$ in the context cache instead of putting in new ones until the frames corresponding to the chunk $i$ are sufficient over time.
More structure details of the IBU are shown in Appendix A.1.

\vspace{-0.2cm}
\subsection{Conversation States Understanding Module}
\vspace{-0.1cm}
In this section, we first analyze the various conversational states of the agent in detail and further design a conversational state understanding module (CSU) to obtain a state guidance for motion generation.

\begin{figure}[t]
   \includegraphics[width=\linewidth]{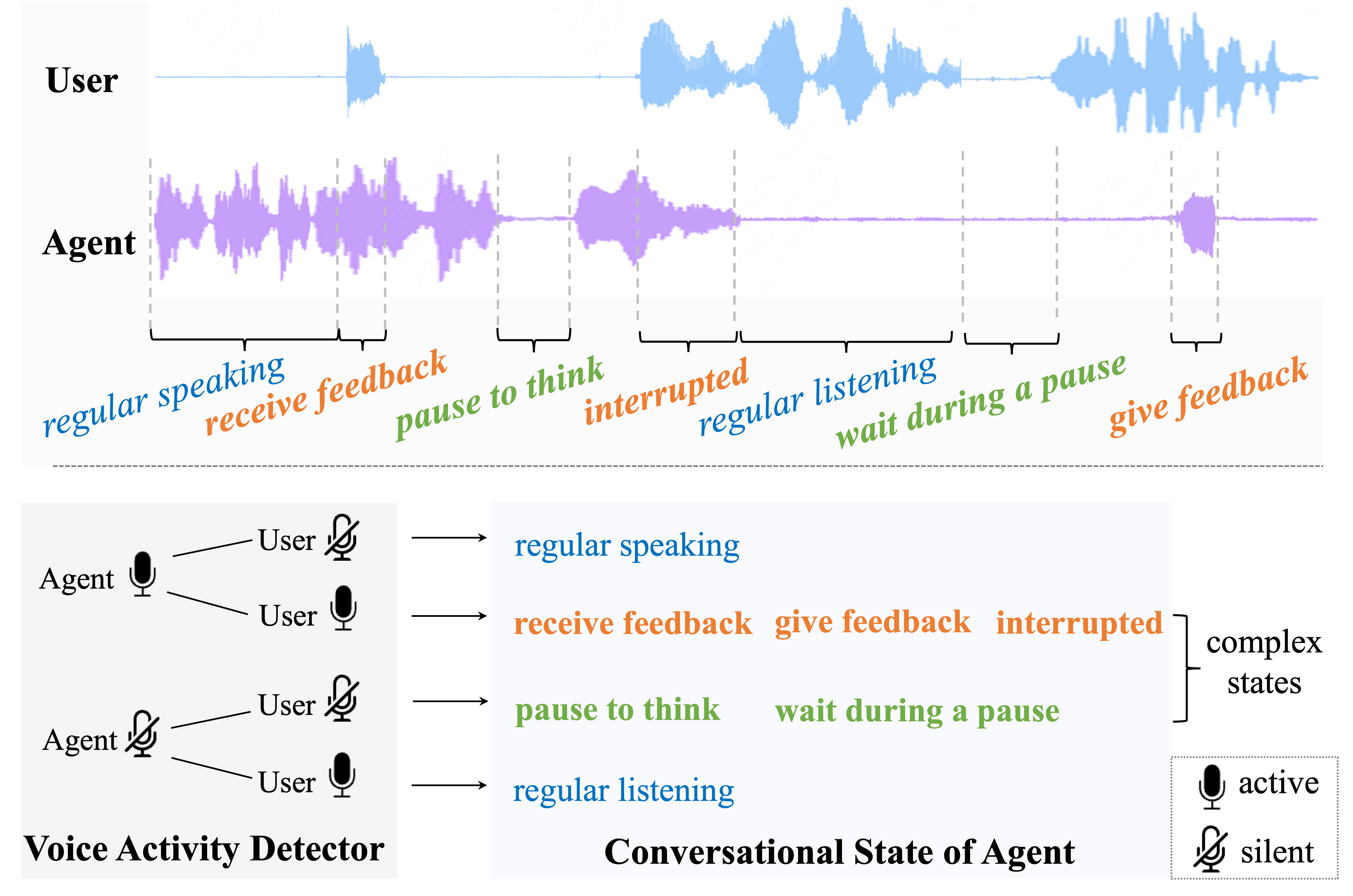}
   \vspace{-0.5cm}
   \caption{Illustration of agent's conversational state. In addition to regular speaking/listening, it also includes many complex states, which are classified according to the voice activity detection.}
   \vspace{-0.5cm}
   \label{fig:state}
\end{figure}

As shown in \cref{fig:state}, besides the regular speaking and listening, the agent encounters more complex states, such as receiving feedback (``wow", ``it's amazing", etc.) while speaking, pausing to think about what to say next, being interrupted, waiting during the other's pause, or giving feedback while listening.
Even with the same audio, the motions differ in different states. For example, given a silent audio, if the state is ``pause to think", it may display a thoughtful expression with its eyes slightly tilted to one side. If the state is ``wait during a pause", it may keep previous motions to patiently wait for the next speech.
Thus, state prediction can guide motions, creating an explicit correlation with motions of similar states to enable more discriminative feature learning, so as to enhance facial expressiveness.

Specifically, we combine audio signals and contextual interaction summary (cis-token) to predict the states classified into seven categories. We first use a pre-trained Voice Activity Detector (VAD)\footnote{https://github.com/wiseman/py-webrtcvad} to predict whether the current frame of agent's and user's audio are silent/active based on a segment of audio. These 0/1 signals can roughly divide the state into four categories, as shown in \cref{fig:state}. Then we embed VAD signals into vectors which together with cis-token constitute the state condition for specific state prediction.

As shown in the upper right part of \cref{fig:main}, we encode the state index $s_{T-1}$ into a vector as the query, let the state condition be the key and the value, and obtain the state latent feature $S_{T}$ through the cross attention mechanism.
The state category index $s_{T}$ is obtained by SoftMax operation on $S_{T}$, and we use the cross-entropy loss to constrain the latent state feature $S_{T}$.
Then we update $s_{T-1}$ to prepare for the next frame prediction.
The latent state feature $S_{T}$ is then input into the following PMP module as a state guidance.

\subsection{Progressive Motion Prediction Module}
After obtaining the cis-token and state features, the PMP module in this section focuses on how to utilize them effectively to generate motions.
First, we predict the coarse-grained motion outline based on the previous motion $M_{T\!-\!1}^a$, conditioned on the updated audio in the historical cache. Subsequently, we preform the fine-grained prediction based on cis-token, state, and the latest audio features, and ensure the inter-frame continuity through the temporal layer. The output is represented as latent $z$.

For the latent $z \!\in\! R^r$ produced by AR process, the final $M_T^a$ is obtained by the sampler with the distribution $M_T^a \sim p(M_T^a \!\mid\! z)$.
Conceptually, for common discretely quantized AR, $M_T^a$ is the index of a codebook with the size $N$, and the sampler is a $N$-classifier which is performed by a Softmax operation with categorical probability distribution, and the cross-entropy is used to calculate the loss.
However, $M^a$ include keypoints coordinates of optical flow within $0\!\sim\!1$, where even small shifts can affect facial motions, and micro-expressions are crucial for conveying information. Thus, compressing these coordinates into $N$ combinations may reduce facial expressiveness, so we continue to model them in a continuous, non-quantized form.

Following \cite{deng2024nova,mar}, to achieve continuous non-quantized AR, we use the diffusion procedure to represent the probability distribution $p(M_T^a\!\mid\!z)$ of the sampler, and the metric loss can be formulated as a denoising criterion:
\begin{equation}
\mathcal{L}(z, M_T^a)=\mathbb{E}_{\varepsilon, t}\left[\left\|\varepsilon-\varepsilon_\theta\left({M_T^a}_t \mid t, z\right)\right\|^2\right]
\end{equation}
where $\varepsilon_\theta$ is the noise estimator to serve as the motion sampler, which is a MLP network parameterized by $\theta$ with AdaLN to inject the condition $z$.
$\varepsilon$ is a Gaussian noise sampled from $\mathcal{N}(\mathbf{0}, \mathbf{I})$, $t$ is a time step. Let $\bar{\alpha}_t$ denote a noise schedule, $\sigma_t$ denote the noise level, then $M_T^a$ is sequentially denoised by :
\begin{equation}
{M_T^a}_{t-1}=\frac{1}{\sqrt{\alpha_t}}\left({M_T^a}_{t}-\frac{1-\alpha_t}{\sqrt{1-\bar{\alpha}_t}} \varepsilon_\theta\left({M_T^a}_{t} \mid t, z\right)\right)+\sigma_t \varepsilon
\end{equation}
The detailed model structures are shown in Appendix A.1.
\section{Experiments}

\begin{figure*}[t]
   \includegraphics[width=\linewidth]{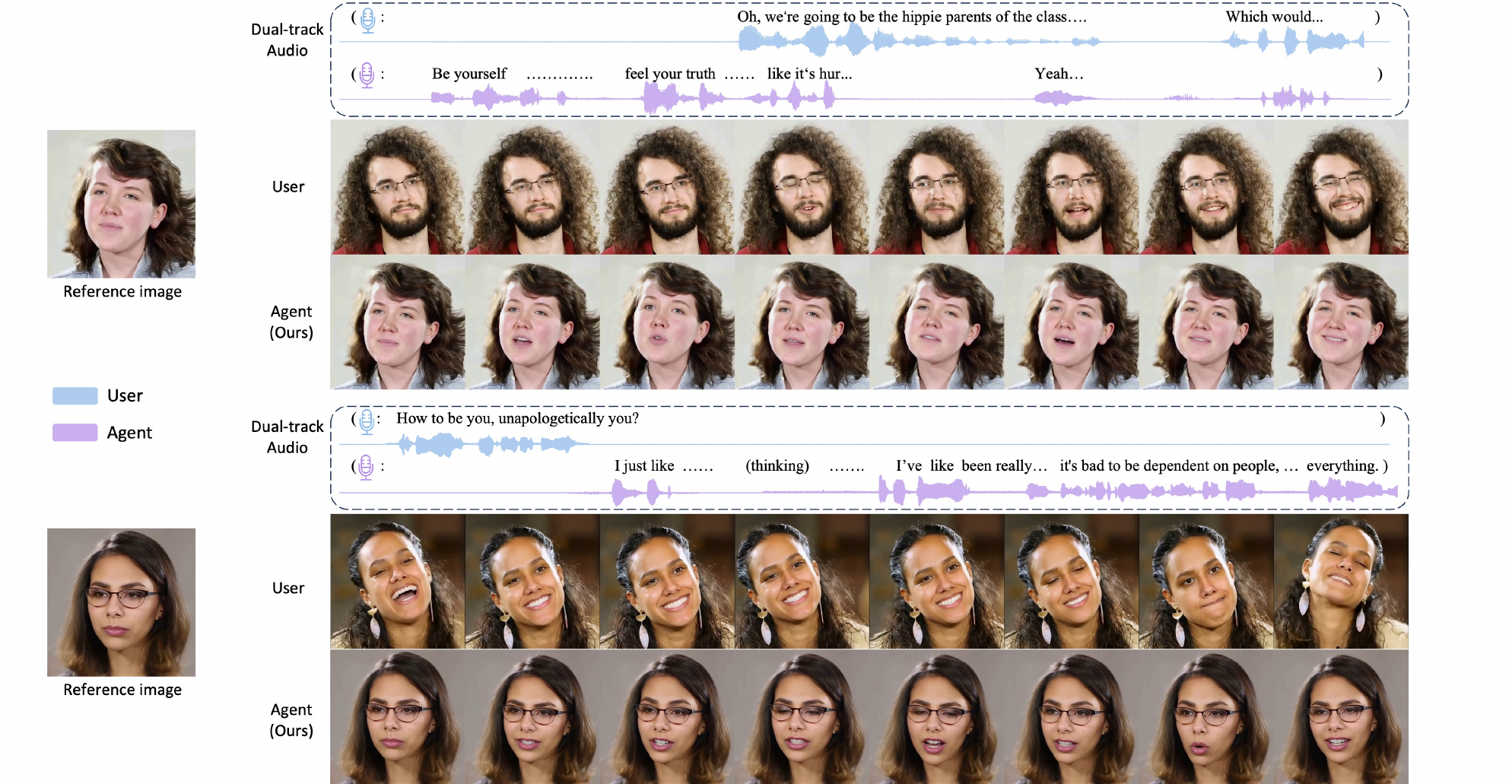}
   \caption{Visual results of our method on RealTalk\cite{realtalk} dataset. The two shown scenarios involves regular speaking/listening, and complex states such as interruptions, feedback, and pauses for thinking.
   Our method can generate highly realistic and natural agent videos in complex conversational scenarios.}
   \vspace{-0.5cm}
   \label{fig:exp_fig1}
\end{figure*}

\subsection{Experimental Settings}
\paragraph{Datasets}
For training data, we first use MultiDialog\cite{park2024let} and ViCo\cite{vico} for training in regular talking and listening scenarios, respectively. Then we use RealTalk\cite{realtalk} for diverse interactive scenario training. In order to ensure data quality, we carry out several-step data cleaning, removing data with fast camera movements and insufficient effective facial regions, ultimately retaining more than 200 hours of data. Then, we extract motions by \cite{liveportrait} and extract audio features from a pretrained feature extractor Wav2Vec2\cite{wav2vec2}. Furthermore, to learn conversational states, we conduct additional state annotations on the training dataset.

\vspace{-0.4cm}
\paragraph{Implementation Details}
We divide frames at 25fps, with each frame corresponding to 40ms. The values are $c\!=\!6$, $w\!=\!512$, $h\!=\!6$. 
The inference fps of the generation process is 31, which meets the real-time requirements.
We use AdamW optimizer \cite{adamw} to train our model, and the base learning rate is set to $1e^{-4}$. The diffusion step during inference is 15. The videos are generated by calculating the optical flow relative to the reference image and warping appearance features using the method in \cite{liveportrait}. More details are shown in Appendix A.2.

\begin{table*}[t]
\footnotesize
\centering

\begin{tabular}{c cc cc cc cc}
\toprule

    Methods & RPCC $\downarrow$ & CSIM $\uparrow$ & SyncScore $\uparrow$ & PSNR $\uparrow$ & SSIM $\uparrow$ & FID $\downarrow$ & SID$\uparrow$ & Var$\uparrow$  \\ \hline
    DIM \cite{dim} & 0.186 & 0.843 & 4.192 & 21.31 & 0.573 & 26.29 & 1.083 & 1.206 
    \\  

    GT   & 0.000 & 0.964 & 7.322 & N/A & 1.000 & 0.000 &  2.902 & 2.514  \\

    Ours & \textbf{0.125} & \textbf{0.901} & \textbf{7.218} & \textbf{29.67} & \textbf{0.827} & \textbf{21.64} & \textbf{2.428} & \textbf{2.397}  \\ \midrule
    w/o long-range context             & 0.140 & 0.885 & 7.128 & 27.09 & 0.803 & 24.68 & 2.381 & 2.326  \\
    w/o conversational state guidance    & 0.127 & 0.892 & 6.973 & 28.14 & 0.816 & 22.96 & 2.263 & 2.174  \\
    w/o visual modality in IBU   & 0.155 & 0.853 & 7.197 & 27.38 & 0.812 & 22.34 & 2.115 & 2.032  \\

\bottomrule
\end{tabular}
\caption{Quantitative comparisons on RealTalk\cite{realtalk}. \textbf{Bold} represents the best. The $\uparrow$ indicates higher is better, the $\downarrow$ indicates lower is better.}
\vspace{-0.2cm}
\label{table:main_table}
\end{table*}

\vspace{-0.1cm}
\subsection{Interactive Head Generation} 
\vspace{-0.1cm}
\subsubsection{Qualitative Results}
\paragraph{Visual Results}
We present the visual results generated by our method in two complex conversation scenarios in Figure \ref{fig:exp_fig1}, which, in addition to regular speaking/listening, also involve complex states such as interruptions, feedback, and pauses for thinking.
Given the reference image, the user's video frames, and the dual-track audio (user's audio and agent's audio), our method can generate realistic and reasonable agent motion for different conversation content. 
Specifically,
in the first example, we can see that the generated agent exhibits natural transitions when interrupted by the user. This can be attributed to our designs of the conversational state and the strong contextual understanding between two parties. Additionally, the agent motions can fluctuate naturally in response to the user's behavior, such as smiling, demonstrating the benefits of incorporating the user's visual information. 
In the second example, when both parties in the conversation are silent (i.e., no speech input), our model employs contextual understanding and state learning to interpret the current pause as an intention to continue speaking,
thereby generating motions that exhibit a ``thinking" expression, resulting in a very natural and smooth conversational performance.

\begin{figure}[t]
    \vspace{-0.2cm}
   \includegraphics[width=\linewidth]{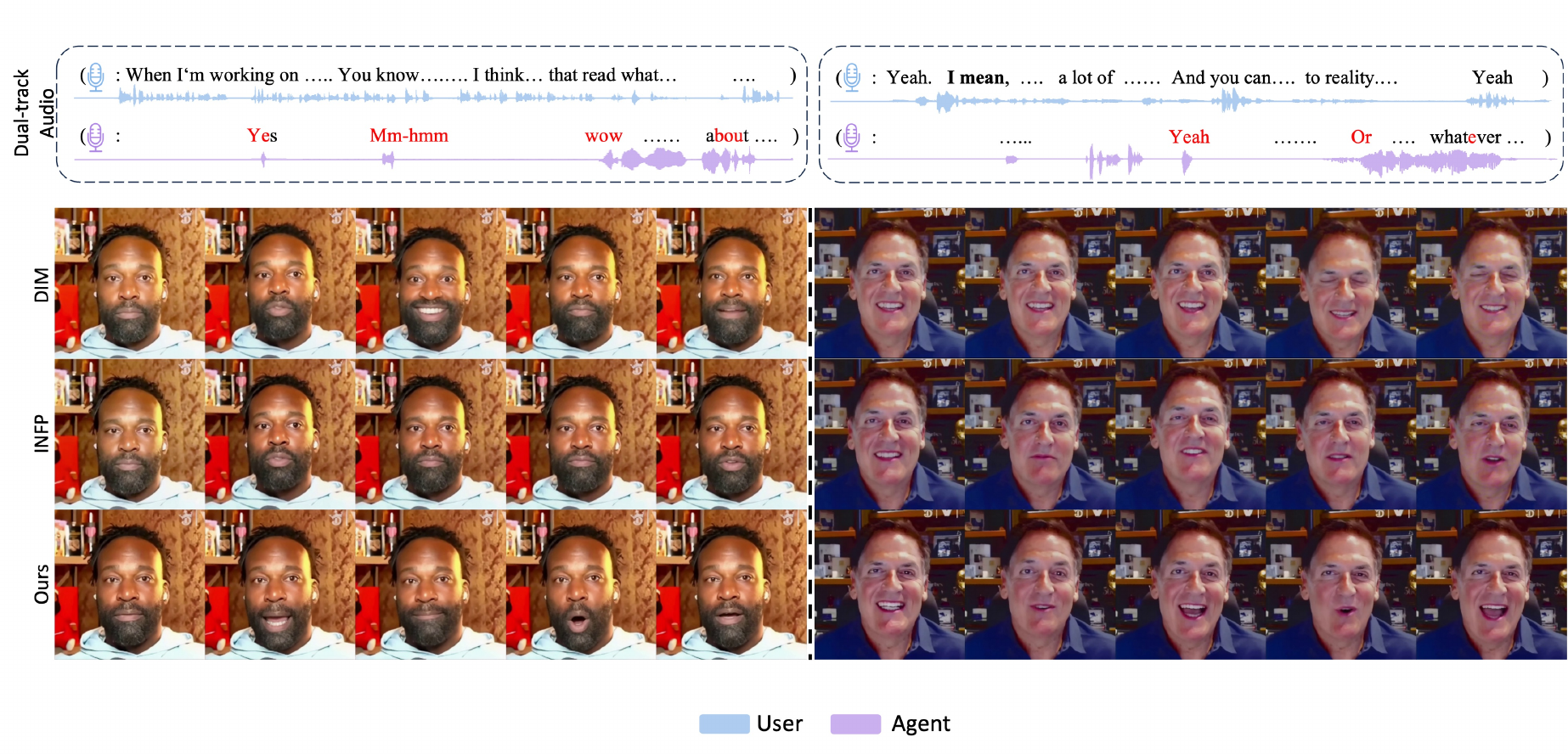}
   \caption{Qualitative comparisons with DIM\cite{dim} and INFP\cite{infp}. The two sample videos are from the DyConv dataset proposed by INFP\cite{infp}, which is not open-sourced, thus lacking ground truth.}
   \vspace{-0.3cm}
   \label{fig:exp_fig2}
\end{figure}

\vspace{-0.5cm}
\paragraph{Comparisons}
In Figure \ref{fig:exp_fig2}, we present qualitative comparisons
with DIM~\cite{dim} and INFP\cite{infp} conditioned on the same reference image and the same audios. 
Since INFP\cite{infp} is not open-source, we directly utilize two videos from the website of INFP\cite{infp} for visual comparison. 
Due to the lack of the ground-truth user videos, we first generate talking head videos based on the given user audio as the user video input to our model, and then generate agent motions. 
It can be seen that the agent videos generated by our method exhibit more realistic and diverse facial expressions in both speaking and listening. For speaking, when pronouncing some specific words (\eg ``Yes", ``Mm-hmm", ``wow", ``about", ``Yeah", ``Or" and ``whatever"), the lip movements generated by ours are more accurate and natural. For listening, our generated agent can display a natural ``curious" expression when the user says ``I mean", demonstrating the advantages of our method in generating highly natural and realistic facial motions compared to other methods. 
In contrast, DIM\cite{dim} suffers from ill-timed and inaccurate responses, and INFP\cite{infp} suffers from unnatural facial expressions (\eg suboptimal lip movements and disordered eye gaze).
More comparisons with DIM are in Appendix C.1. The full generated videos can be found in our supplementary video.

\subsubsection{Quantitative Results} 
\paragraph{Metrics}
We evaluate our proposed methods from five aspects, including motion synchronization between conversation partners, identity preservation, lip-sync performance, video realism and motion diversity. For motion synchronization, we utilize Residual Pearson Correlation Coefficient (RPCC) to measure the correlation between the user motions and the agent motions. For identity preservation, we employ CSIM to calculate the the cosine similarity of identity features between the reference image and the generated video. For lip-sync performance, we utilize SyncScore \cite{wav2lip} to measure the lip-sync accuracy in generated videos. For video realism, we utilize Peak Signal-to-Noise Ratio (PSNR) and Structural Similarity (SSIM) to measure the pixel-level similarity between the generated video frames and ground-truth video frames, and adopt Frechet Inception
Distance (FID) to measure the feature-level difference between them. For motion diversity, we employ SI\cite{l2l} for Diversity (SID) and Variance (Var).

\vspace{-0.5cm}
\paragraph{Comparisons}
As our method is not a 3DMM-based method, we first extract 70-dim 3DMM coefficients (64-dim for expression and 6-dim for pose) from our generated videos by a pretrained deep 3D face reconstruction method\cite{deep3d}, and then use them for the evaluation of RPCC, SID and Var. Since INFP\cite{infp} is not open-source, we only retrained DIM\cite{dim} on our dataset and present the quantitative comparisons between our method and DIM\cite{dim} in Table \ref{table:main_table}. It can be seen that our method outperforms DIM\cite{dim} across multiple aspects. For example, our method achieves the lowest RPCC, indicating superior motion synchronization ability, which can be attributed to the effective modeling and understanding of interactive behavior between the user and the agent. Additionally, the SyncScore \cite{wav2lip} of our method outperforms DIM\cite{dim} by a large margin, which benefits from the introduction of the conversational state as well as the well-designed PMP module. Furthermore, compared to DIM\cite{dim}, our proposed method exhibits higher SID and Var, showing great advantages in terms of motion diversity. For video realism, our method has the best performance in PSNR, SSIM and FID, which demonstrates that the motions generated by our model are closest to the ground-truth motions. Although identity preservation is not our main focus,
we still achieve some performance improvement in CSIM.


\begin{table}[t]
\footnotesize
\centering
\setlength\tabcolsep{0.15cm}
\begin{tabular}{c c c c c c c c c}
\toprule
    \multirow{2}{*}{Methods} &\multicolumn{2}{c}{FD $\downarrow$} &   \multicolumn{2}{c}{RPCC $\downarrow$} & 
  \multicolumn{2}{c}{SID$\uparrow$} & \multicolumn{2}{c}{Var$\uparrow$} \\ 

  \cmidrule(lr){2-3} \cmidrule(lr){4-5} \cmidrule(lr){6-7}
  \cmidrule(lr){8-9}

    &  exp  &  pose &  exp  &  pose &  exp  &  pose &  exp  &  pose \\ \hline
    
    \hline
    RLHG$^{\ast}$ \cite{vico} & 39.02 & 0.07 & 0.08 & 0.02 & 3.62 & 3.17 & 1.52 & 0.02 \\
    L2L$^{\ast}$ \cite{l2l} &   33.93 & \textbf{0.06} & 0.06 & 0.08 & 2.77 & 2.66 & 0.83 & 0.02 \\
    DIM$^{\ast}$ \cite{dim} &  23.88 &  \textbf{0.06} &  0.06 & 0.03 & 3.71 & 2.35 & 1.53 & 0.02\\
    INFP$^{\ast}$ \cite{infp} & 18.63 & 0.07 & - & - & 4.78 & 3.92 & 2.83 & \textbf{0.18} \\
    Ours &  \textbf{18.39} &  \textbf{0.06} &  \textbf{0.05} &  \textbf{0.01}  &  \textbf{4.82} &  \textbf{3.94} &  \textbf{2.91}  &  0.17 \\

\bottomrule
\end{tabular}
\vspace{-0.2cm}
\caption{Quantitative results with state-of-the-art listening head generation methods on ViCo\cite{vico} dataset. $^{\ast}$ denotes the results are inherited from DIM\cite{dim} and INFP\cite{infp}.}
\label{table:listening_head_cmp}
\end{table}
\begin{table}[t]
\vspace{-0.1cm}
\setlength{\tabcolsep}{4pt}
\footnotesize
\centering
\begin{tabular}{c c c c c c}
\toprule
    Method & PSNR $\uparrow$ & SSIM $\uparrow$ & FID $\downarrow$ & CSIM $\uparrow$ & SyncScore $\uparrow$ \\ \hline
    SadTalker\cite{zhang2023sadtalker}  &  25.65  &  0.786  &  23.46  &  0.821 & 6.792\\
    Hallo\cite{xu2024hallo} &  28.32  &  0.801  &  21.77 &  0.860 & 7.116\\
    EchoMimic\cite{chen2024echomimic} &  \textbf{28.65}  &  0.805  &  20.91 &  0.862 & 7.382\\
    Ours &  28.63  &  \textbf{0.806}  &  \textbf{18.32}  & \textbf{0.876} & \textbf{7.424} \\
      
\bottomrule
\end{tabular}
\vspace{-0.15cm}
\caption{Quantitative results with state-of-the-art talking head generation methods on HDTF\cite{hdtf} dataset.}
\vspace{-0.4cm}
\label{table:talking_head_cmp}
\end{table}

\vspace{-0.1cm}
\subsection{Single-role Head Generation} 
In addition to dyadic interactive head generation, our method can be directly applied to single-role head generation, such as listening head generation and talking head generation, without additional fine-tuning. To further validate the superiority of our proposed method, in the following, we will compare our method with the state-of-the-art listening head generation and talking head generation methods.

\vspace{-0.1cm}
\subsubsection{Listening Head Generation} 
Following DIM\cite{dim}, we utilize Frechet Distance (FD) to evaluate motion realism of the generated listener, RPCC for motion synchronization, and SID and Var for motion diversity. We present quantitative comparisions with the state-of-the-art listening head generation methods (\eg RLHG\cite{vico}, L2L\cite{l2l}, DIM\cite{dim} and INFP\cite{infp}) on ViCo\cite{vico} dataset in Table \ref{table:listening_head_cmp}. It can be obviously seen that our proposed method has the lowest FD values for both expression and pose, indicating that our method can generate highly lifelike facial and head movements, which validates the effectiveness of our structural design. Furthermore, the RPCC of our method outperforms other methods, which justifies the superiority of our model in dyadic conversation modeling. For SID and Var, our method also shows a great advantage compared to other SOTA methods, which further demonstrates that the motions generated by our method is more natural and realistic. For visual comparisons, please refer to our Appendix C.3 and supplementary video.

\vspace{-0.1cm}
\subsubsection{Talking Head Generation} 
We choose several widely used metrics in talking head generation for evaluation (\eg PSNR, SSIM, FID, CSIM and SyncScore \cite{wav2lip}), and we compare our method with SadTalker \cite{zhang2023sadtalker}, Hallo \cite{xu2024hallo} and EchoMimic \cite{chen2024echomimic}. Following INFP \cite{infp}, we randomly selected 50 videos from HDTF \cite{hdtf} dataset as the test data. As shown in Table \ref{table:talking_head_cmp}, our method achieves the best performance in metrics like SyncScore \cite{wav2lip}, CSIM and FID, showing great superiority in lip synchronization, identity preservation and feature-level video quality. Moreover, although EchoMimic \cite{chen2024echomimic} is grounded in  Stable Diffusion (SD) v1.5 \cite{ldm}, our method still has comparable performance with it on metrics related to image quality (\eg PSNR and SSIM). Apart from quantitative comparison, we also present detailed visual comparisons in Appendix C.2 and the supplementary video.


\begin{figure}[t]
   \includegraphics[width=\linewidth]{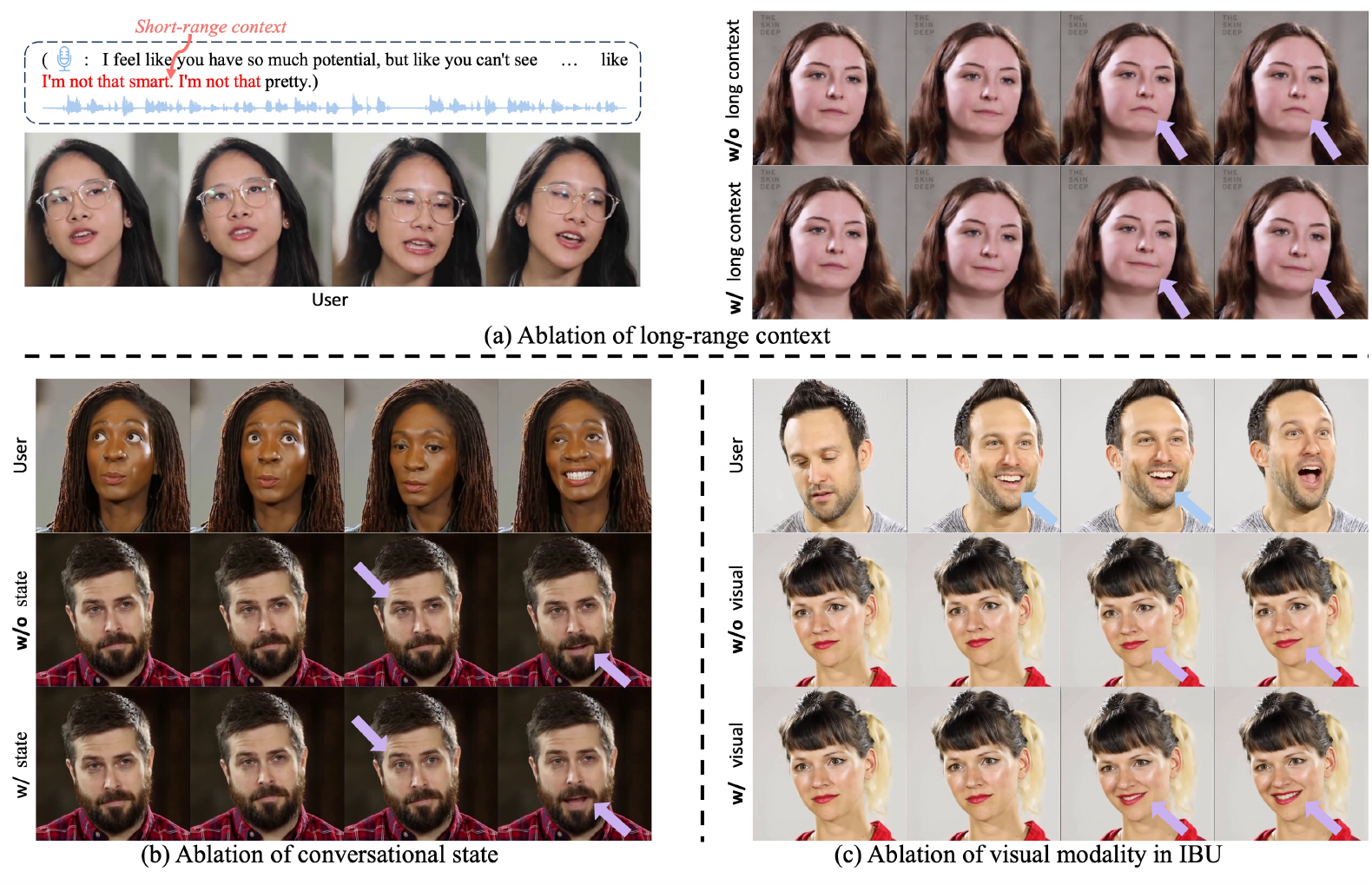}
   \caption{Ablation study. (a) Ablation of long-range context. (b) Ablation of conversational state. (c) Ablation of visual modality in the IBU module.}
   \vspace{-0.6cm}
   \label{fig:exp_ablation}
\end{figure}

\vspace{-0.1cm}
\subsection{Ablation Study}
In this section, we present ablation results of long-range context, conversational state and visual modality in IBU. Additional ablation for continuous AR modeling and bidirectional-integrated learning are in Appendix B.1.

\vspace{-0.4cm}
\paragraph{Long-range Context}
We utilize long-range context in IBU module for more comprehensive contextual information understanding. To validate its effectiveness, we set the capacity of context cache $w$ as 4 in IBU and only consider short-range context (with a length of 1s, refer to \cite{wang2023agentavatar}). The quantitative results are shown in Table \ref{table:main_table}. Without long-range context, the RPCC worsened greatly, which validates long-range context is crucial for semantic understanding and thus has a significant effect on motion synchronization between two parties in conversation. The qualitative results are shown in Figure \ref{fig:exp_ablation} (a). It can be seen that without long-range context, the model may only consider the short-range context (\eg ``I'm not that smart...") and misunderstand the contextual information. This can lead to incorrect feedback, such as mistakenly shifting the attitude from affirmation (\eg smile) to disapproval (\eg downturned mouth).

\vspace{-0.4cm}
\paragraph{Conversational State Guidance}
To validate the effectiveness of conversation state, we remove the CSU module in our framework. As shown in Table \ref{table:main_table}, without conversation state, the SyncScore, FID, PSNR and SSIM are all worsened, which validates that the conversation state is capable of providing a strong guidance for motion prediction and making the generated motion more realistic and closer to the ground truth. In Figure \ref{fig:exp_ablation} (b), we can also see that the conversation state can greatly enhance facial expressiveness (\eg wide-open eyes, more precise lip motion).

\vspace{-0.4cm}
\paragraph{Visual Modality}
We remove the visual modality in the IBU module and only consider audio features. The ablation results are shown in Table \ref{table:main_table}. It can be seen that the RPCC increases significantly, indicating suboptimal motion synchronization without visual modality. The similar conclusion can also be reflected in Figure \ref{fig:exp_ablation} (c), with the incorporation of visual modality, the agent’s facial motions can fluctuate naturally with the user’s motions (\eg smile).

\section{Conclusion}
In this paper, we propose a frame-wise framework called ARIG based on continuous non-VQ autoregression to realize real-time and more realistic interactive head generation.
In order to learn long-range contextual behavior, we design the IBU module which summarizes short-chunk audio-visual behaviors through Bidirectional-Integrated learning and then aggregates them to achieve long-range understanding.
To understand complex states of real-world conversations, CSU utilizes voice activity signals and long-range context to generate the state feature. 
In PMP, we progressively predict motions and achieve more accurate continuous motion generation through diffusion probability modeling in the DiffusionMLP.
Comprehensive experiments validate the superiority of our proposed method.

{
    \small
    \bibliographystyle{ieeenat_fullname}
    \bibliography{main}

\begin{thebibliography}{39}
\providecommand{\natexlab}[1]{#1}
\providecommand{\url}[1]{\texttt{#1}}
\expandafter\ifx\csname urlstyle\endcsname\relax
  \providecommand{\doi}[1]{doi: #1}\else
  \providecommand{\doi}{doi: \begingroup \urlstyle{rm}\Url}\fi

\bibitem[Baevski et~al.(2020)Baevski, Zhou, Mohamed, and Auli]{wav2vec2}
Alexei Baevski, Yuhao Zhou, Abdelrahman Mohamed, and Michael Auli.
\newblock wav2vec 2.0: A framework for self-supervised learning of speech representations.
\newblock \emph{Advances in neural information processing systems}, 33:\penalty0 12449--12460, 2020.

\bibitem[Chen et~al.(2024)Chen, Cao, Chen, Li, and Ma]{chen2024echomimic}
Zhiyuan Chen, Jiajiong Cao, Zhiquan Chen, Yuming Li, and Chenguang Ma.
\newblock Echomimic: Lifelike audio-driven portrait animations through editable landmark conditions.
\newblock \emph{arXiv preprint arXiv:2407.08136}, 2024.

\bibitem[Dehghani et~al.(2023)Dehghani, Djolonga, Mustafa, Padlewski, Heek, Gilmer, Steiner, Caron, Geirhos, Alabdulmohsin, et~al.]{dehghani2023scaling}
Mostafa Dehghani, Josip Djolonga, Basil Mustafa, Piotr Padlewski, Jonathan Heek, Justin Gilmer, Andreas~Peter Steiner, Mathilde Caron, Robert Geirhos, Ibrahim Alabdulmohsin, et~al.
\newblock Scaling vision transformers to 22 billion parameters.
\newblock In \emph{International Conference on Machine Learning}, pages 7480--7512. PMLR, 2023.

\bibitem[Deng et~al.(2024)Deng, Pan, Diao, Luo, Cui, Lu, Shan, Qi, and Wang]{deng2024nova}
Haoge Deng, Ting Pan, Haiwen Diao, Zhengxiong Luo, Yufeng Cui, Huchuan Lu, Shiguang Shan, Yonggang Qi, and Xinlong Wang.
\newblock Autoregressive video generation without vector quantization.
\newblock \emph{arXiv preprint arXiv:2412.14169}, 2024.

\bibitem[Deng et~al.(2019)Deng, Yang, Xu, Chen, Jia, and Tong]{deep3d}
Yu Deng, Jiaolong Yang, Sicheng Xu, Dong Chen, Yunde Jia, and Xin Tong.
\newblock Accurate 3d face reconstruction with weakly-supervised learning: From single image to image set.
\newblock In \emph{Proceedings of the IEEE/CVF conference on computer vision and pattern recognition workshops}, pages 0--0, 2019.

\bibitem[Esser et~al.(2024)Esser, Kulal, Blattmann, Entezari, M{\"u}ller, Saini, Levi, Lorenz, Sauer, Boesel, et~al.]{mmdit}
Patrick Esser, Sumith Kulal, Andreas Blattmann, Rahim Entezari, Jonas M{\"u}ller, Harry Saini, Yam Levi, Dominik Lorenz, Axel Sauer, Frederic Boesel, et~al.
\newblock Scaling rectified flow transformers for high-resolution image synthesis.
\newblock In \emph{Forty-first international conference on machine learning}, 2024.

\bibitem[Geng et~al.(2023)Geng, Teotia, Tendulkar, Menon, and Vondrick]{realtalk}
Scott Geng, Revant Teotia, Purva Tendulkar, Sachit Menon, and Carl Vondrick.
\newblock Affective faces for goal-driven dyadic communication.
\newblock \emph{arXiv preprint arXiv:2301.10939}, 2023.

\bibitem[Guo et~al.(2024)Guo, Zhang, Liu, Zhong, Zhang, Wan, and Zhang]{liveportrait}
Jianzhu Guo, Dingyun Zhang, Xiaoqiang Liu, Zhizhou Zhong, Yuan Zhang, Pengfei Wan, and Di Zhang.
\newblock Liveportrait: Efficient portrait animation with stitching and retargeting control.
\newblock \emph{arXiv preprint arXiv:2407.03168}, 2024.

\bibitem[Huang et~al.(2022)Huang, Huang, and Zhou]{pchg}
Ailin Huang, Zhewei Huang, and Shuchang Zhou.
\newblock Perceptual conversational head generation with regularized driver and enhanced renderer.
\newblock In \emph{Proceedings of the 30th ACM International Conference on Multimedia (MM'22)}, 2022.

\bibitem[Huang et~al.(2024)Huang, Ho, Qin, Shi, and Komura]{huang2024interact}
Yinghao Huang, Leo Ho, Dafei Qin, Mingyi Shi, and Taku Komura.
\newblock Interact: Capture and modelling of realistic, expressive and interactive activities between two persons in daily scenarios.
\newblock \emph{arXiv preprint arXiv:2405.11690}, 2024.

\bibitem[Jiang et~al.(2024)Jiang, Liang, Yang, Lin, Zhong, and Zheng]{jiang2024loopy}
Jianwen Jiang, Chao Liang, Jiaqi Yang, Gaojie Lin, Tianyun Zhong, and Yanbo Zheng.
\newblock Loopy: Taming audio-driven portrait avatar with long-term motion dependency.
\newblock \emph{arXiv preprint arXiv:2409.02634}, 2024.

\bibitem[Kolesnikov et~al.(2022)Kolesnikov, Susano~Pinto, Beyer, Zhai, Harmsen, and Houlsby]{kolesnikov2022uvim}
Alexander Kolesnikov, Andr{\'e} Susano~Pinto, Lucas Beyer, Xiaohua Zhai, Jeremiah Harmsen, and Neil Houlsby.
\newblock Uvim: A unified modeling approach for vision with learned guiding codes.
\newblock \emph{Advances in Neural Information Processing Systems}, 35:\penalty0 26295--26308, 2022.

\bibitem[Li et~al.(2025)Li, Tian, Li, Deng, and He]{mar}
Tianhong Li, Yonglong Tian, He Li, Mingyang Deng, and Kaiming He.
\newblock Autoregressive image generation without vector quantization.
\newblock \emph{Advances in Neural Information Processing Systems}, 37:\penalty0 56424--56445, 2025.

\bibitem[Liu et~al.(2023)Liu, Wang, Fu, Chai, Yu, Dai, and Han]{mfr}
Jin Liu, Xi Wang, Xiaomeng Fu, Yesheng Chai, Cai Yu, Jiao Dai, and Jizhong Han.
\newblock Mfr-net: Multi-faceted responsive listening head generation via denoising diffusion model.
\newblock In \emph{Proceedings of the 31th ACM International Conference on Multimedia (MM'23)}, 2023.

\bibitem[Liu et~al.(2024)Liu, Guo, Zhen, Li, Ao, and Yan]{liu2024customlistener}
Xi Liu, Ying Guo, Cheng Zhen, Tong Li, Yingying Ao, and Pengfei Yan.
\newblock Customlistener: Text-guided responsive interaction for user-friendly listening head generation.
\newblock In \emph{Proceedings of the IEEE/CVF Conference on Computer Vision and Pattern Recognition}, pages 2415--2424, 2024.

\bibitem[Loshchilov and Hutter(2017)]{adamw}
Ilya Loshchilov and Frank Hutter.
\newblock Decoupled weight decay regularization.
\newblock \emph{arXiv preprint arXiv:1711.05101}, 2017.

\bibitem[Ng et~al.(2022)Ng, Joo, Hu, Li, Darrell, Kanazawa, and Ginosar]{l2l}
Evonne Ng, Hanbyul Joo, Liwen Hu, Hao Li, Trevor Darrell, Angjoo Kanazawa, and Shiry Ginosar.
\newblock Learning to listen: Modeling non-deterministic dyadic facial motion.
\newblock In \emph{Proceedings of the IEEE/CVF Conference on Computer Vision and Pattern Recognition (CVPR)}, pages 20395--20405, 2022.

\bibitem[Ng et~al.(2024)Ng, Romero, Bagautdinov, Bai, Darrell, Kanazawa, and Richard]{ng2024audio}
Evonne Ng, Javier Romero, Timur Bagautdinov, Shaojie Bai, Trevor Darrell, Angjoo Kanazawa, and Alexander Richard.
\newblock From audio to photoreal embodiment: Synthesizing humans in conversations.
\newblock In \emph{Proceedings of the IEEE/CVF Conference on Computer Vision and Pattern Recognition}, pages 1001--1010, 2024.

\bibitem[Park et~al.(2024)Park, Kim, Rha, Kim, Hong, Yeo, and Ro]{park2024let}
Se~Jin Park, Chae~Won Kim, Hyeongseop Rha, Minsu Kim, Joanna Hong, Jeong~Hun Yeo, and Yong~Man Ro.
\newblock Let's go real talk: Spoken dialogue model for face-to-face conversation.
\newblock \emph{arXiv preprint arXiv:2406.07867}, 2024.

\bibitem[Peebles and Xie(2023)]{dit}
William Peebles and Saining Xie.
\newblock Scalable diffusion models with transformers.
\newblock In \emph{Proceedings of the IEEE/CVF international conference on computer vision}, pages 4195--4205, 2023.

\bibitem[Prajwal et~al.(2020)Prajwal, Mukhopadhyay, Namboodiri, and Jawahar]{wav2lip}
KR Prajwal, Rudrabha Mukhopadhyay, Vinay~P Namboodiri, and CV Jawahar.
\newblock A lip sync expert is all you need for speech to lip generation in the wild.
\newblock In \emph{Proceedings of the 28th ACM international conference on multimedia}, pages 484--492, 2020.

\bibitem[Ramesh et~al.(2021)Ramesh, Pavlov, Goh, Gray, Voss, Radford, Chen, and Sutskever]{ramesh2021zero}
Aditya Ramesh, Mikhail Pavlov, Gabriel Goh, Scott Gray, Chelsea Voss, Alec Radford, Mark Chen, and Ilya Sutskever.
\newblock Zero-shot text-to-image generation.
\newblock In \emph{International conference on machine learning}, pages 8821--8831. Pmlr, 2021.

\bibitem[Rombach et~al.(2022)Rombach, Blattmann, Lorenz, Esser, and Ommer]{ldm}
Robin Rombach, Andreas Blattmann, Dominik Lorenz, Patrick Esser, and Bj{\"o}rn Ommer.
\newblock High-resolution image synthesis with latent diffusion models.
\newblock In \emph{Proceedings of the IEEE/CVF conference on computer vision and pattern recognition}, pages 10684--10695, 2022.

\bibitem[Song et~al.(2023)Song, Yin, Jin, Dong, and Xu]{ELP}
Luchuan Song, Guojun Yin, Zhenchao Jin, Xiaoyi Dong, and Chenliang Xu.
\newblock Emotional listener portrait: Neural listener head generation with emotion.
\newblock In \emph{Proceedings of the IEEE/CVF International Conference on Computer Vision (ICCV)}, pages 20839--20849, 2023.

\bibitem[Sun et~al.(2024)Sun, Xu, Jiang, Liu, Sun, and Huang]{sun2024beyond}
Mingze Sun, Chao Xu, Xinyu Jiang, Yang Liu, Baigui Sun, and Ruqi Huang.
\newblock Beyond talking--generating holistic 3d human dyadic motion for communication.
\newblock \emph{International Journal of Computer Vision}, pages 1--17, 2024.

\bibitem[Tian et~al.(2024)Tian, Wang, Zhang, and Bo]{tian2024emo}
Linrui Tian, Qi Wang, Bang Zhang, and Liefeng Bo.
\newblock Emo: Emote portrait alive generating expressive portrait videos with audio2video diffusion model under weak conditions.
\newblock In \emph{European Conference on Computer Vision}, pages 244--260. Springer, 2024.

\bibitem[Tran et~al.(2024)Tran, Chang, Siniukov, and Soleymani]{dim}
Minh Tran, Di Chang, Maksim Siniukov, and Mohammad Soleymani.
\newblock Dim: Dyadic interaction modeling for social behavior generation.
\newblock In \emph{European Conference on Computer Vision}, pages 484--503. Springer, 2024.

\bibitem[Van Den~Oord et~al.(2017)Van Den~Oord, Vinyals, et~al.]{van2017neural}
Aaron Van Den~Oord, Oriol Vinyals, et~al.
\newblock Neural discrete representation learning.
\newblock \emph{Advances in neural information processing systems}, 30, 2017.

\bibitem[Wang et~al.(2023)Wang, Dai, Deng, and Wang]{wang2023agentavatar}
Duomin Wang, Bin Dai, Yu Deng, and Baoyuan Wang.
\newblock Agentavatar: Disentangling planning, driving and rendering for photorealistic avatar agents.
\newblock \emph{arXiv preprint arXiv:2311.17465}, 2023.

\bibitem[Xu et~al.(2023)Xu, Luo, Xie, Shen, Liu, Liu, Gunes, and Song]{mrecgen}
Jiaqi Xu, Cheng Luo, Weicheng Xie, Linlin Shen, Xiaofeng Liu, Lu Liu, Hatice Gunes, and Siyang Song.
\newblock Mrecgen: Multimodal appropriate reaction generator.
\newblock \emph{arXiv preprint arXiv:2307.02609}, 2023.

\bibitem[Xu et~al.(2024)Xu, Li, Su, Shang, Zhang, Liu, Wang, Yao, and Zhu]{xu2024hallo}
Mingwang Xu, Hui Li, Qingkun Su, Hanlin Shang, Liwei Zhang, Ce Liu, Jingdong Wang, Yao Yao, and Siyu Zhu.
\newblock Hallo: Hierarchical audio-driven visual synthesis for portrait image animation.
\newblock \emph{arXiv preprint arXiv:2406.08801}, 2024.

\bibitem[Yan et~al.(2024)Yan, Zhou, Wang, Gao, and Yang]{dialoguenerf}
Yichao Yan, Zanwei Zhou, Zi Wang, Jingnan Gao, and Xiaokang Yang.
\newblock Dialoguenerf: Towards realistic avatar face-to-face conversation video generation.
\newblock \emph{Visual Intelligence}, 2\penalty0 (1):\penalty0 24, 2024.

\bibitem[Zhang et~al.(2023)Zhang, Cun, Wang, Zhang, Shen, Guo, Shan, and Wang]{zhang2023sadtalker}
Wenxuan Zhang, Xiaodong Cun, Xuan Wang, Yong Zhang, Xi Shen, Yu Guo, Ying Shan, and Fei Wang.
\newblock Sadtalker: Learning realistic 3d motion coefficients for stylized audio-driven single image talking face animation.
\newblock In \emph{Proceedings of the IEEE/CVF conference on computer vision and pattern recognition}, pages 8652--8661, 2023.

\bibitem[Zhang et~al.(2021)Zhang, Li, Ding, and Fan]{hdtf}
Zhimeng Zhang, Lincheng Li, Yu Ding, and Changjie Fan.
\newblock Flow-guided one-shot talking face generation with a high-resolution audio-visual dataset.
\newblock In \emph{Proceedings of the IEEE/CVF Conference on Computer Vision and Pattern Recognition}, pages 3661--3670, 2021.

\bibitem[Zheng et~al.(2024)Zheng, Zhang, Guo, Pan, Tan, Lu, Tang, An, and Yan]{zheng2024memo}
Longtao Zheng, Yifan Zhang, Hanzhong Guo, Jiachun Pan, Zhenxiong Tan, Jiahao Lu, Chuanxin Tang, Bo An, and Shuicheng Yan.
\newblock Memo: Memory-guided diffusion for expressive talking video generation.
\newblock \emph{arXiv preprint arXiv:2412.04448}, 2024.

\bibitem[Zhou et~al.(2022)Zhou, Bai, Zhang, Yao, Zhao, and Mei]{vico}
Mohan Zhou, Yalong Bai, Wei Zhang, Ting Yao, Tiejun Zhao, and Tao Mei.
\newblock Responsive listening head generation: A benchmark dataset and baseline.
\newblock In \emph{Proceedings of the European conference on computer vision (ECCV)}, 2022.

\bibitem[Zhou et~al.(2023{\natexlab{a}})Zhou, Bai, Zhang, Yao, and Zhao]{vicox}
Mohan Zhou, Yalong Bai, Wei Zhang, Ting Yao, and Tiejun Zhao.
\newblock Interactive conversational head generation.
\newblock \emph{arXiv preprint arXiv:2307.02090}, 2023{\natexlab{a}}.

\bibitem[Zhou et~al.(2023{\natexlab{b}})Zhou, Wan, and Wang]{zhou2023unified}
Zixiang Zhou, Yu Wan, and Baoyuan Wang.
\newblock A unified framework for multimodal, multi-part human motion synthesis.
\newblock \emph{arXiv preprint arXiv:2311.16471}, 2023{\natexlab{b}}.

\bibitem[Zhu et~al.(2024)Zhu, Zhang, Rong, Hu, Liang, and Ge]{infp}
Yongming Zhu, Longhao Zhang, Zhengkun Rong, Tianshu Hu, Shuang Liang, and Zhipeng Ge.
\newblock Infp: Audio-driven interactive head generation in dyadic conversations.
\newblock \emph{arXiv preprint arXiv:2412.04037}, 2024.

\end{thebibliography}
}

\newpage
\setcounter{section}{0}
\renewcommand\thesection{\Alph{section}}

\section{Implementation Details}
\subsection{Network Details}
\begin{figure}[t]
   \includegraphics[width=\linewidth]{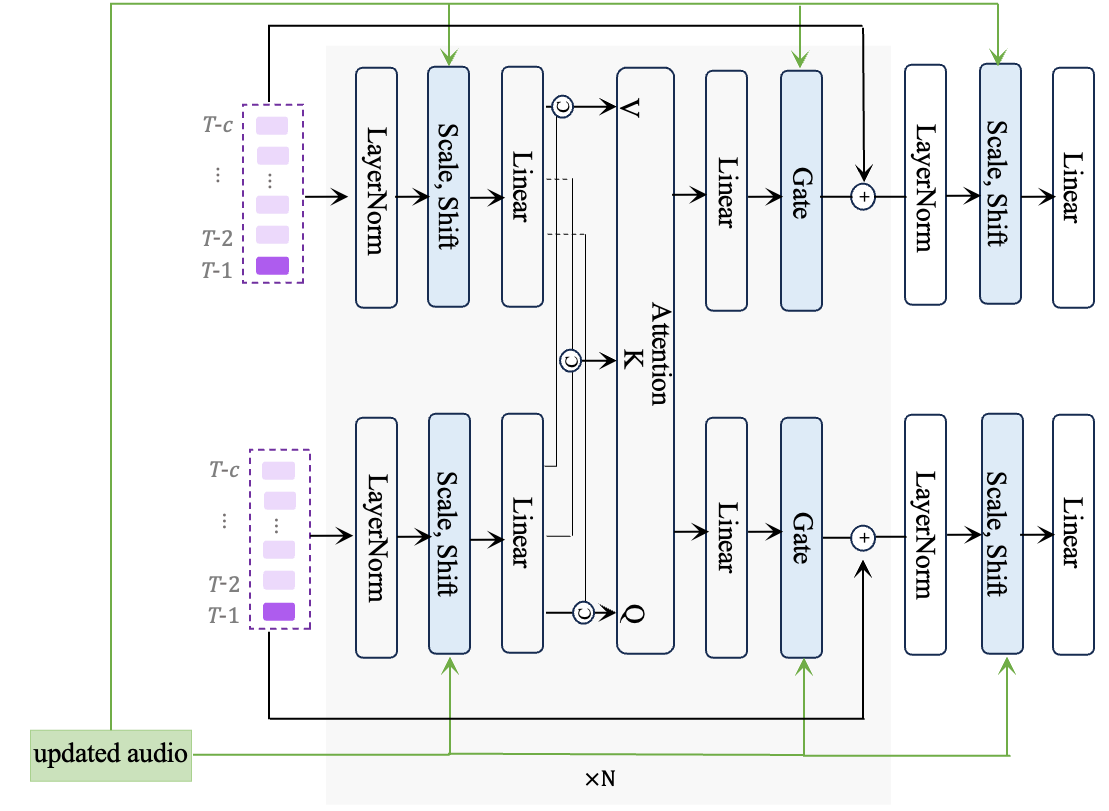}
   \vspace{-0.5cm}
   \caption{The structure of the Bidirectional-learning.}
   \label{fig:dual}
\end{figure}

Due to the page limitation, we show network details of modules in this Appendix.
\vspace{-0.4cm}
\paragraph{Bidirectional-learning} In the bidirectional learning block, we treat the audio-visual behavior of the agent and the user as two independent individuals and use independent model parameters to first understand their own behavior and exchange information through shared attention. 
This independent understanding is conditioned on the updated audio, which is injected by the modulation block. 
We embed the audio in 512-dim and, as adaLN in DiT\cite{dit}, we compute the scale, shift, and gate parameters in the modulation. The depth of the bidirectional learning block is 2. The detailed block structure is shown in \cref{fig:dual}.

\begin{figure}[t]
   \includegraphics[width=\linewidth]{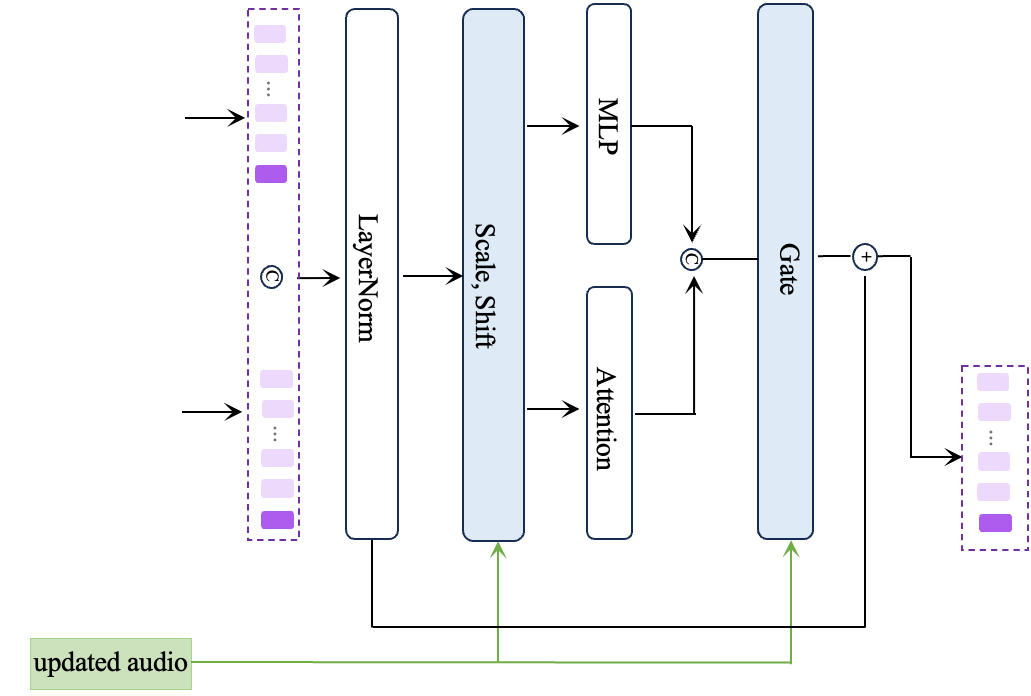}
   \vspace{-0.5cm}
   \caption{The structure of the Integrated-learning.}
   \label{fig:single}
\end{figure}

\vspace{-0.4cm}
\paragraph{Integrated-learning} In the integrated learning stage, we concatenate the outputs of both parties in the Bidirectional-learning to perform unified learning. We also inject the update audio and perform parallel learning of attention and MLP to improve efficiency. Finally, we extract the information of the agent part as the interaction summary. The detailed structure is shown in \cref{fig:single}.

\begin{figure}[t]
   \includegraphics[width=\linewidth]{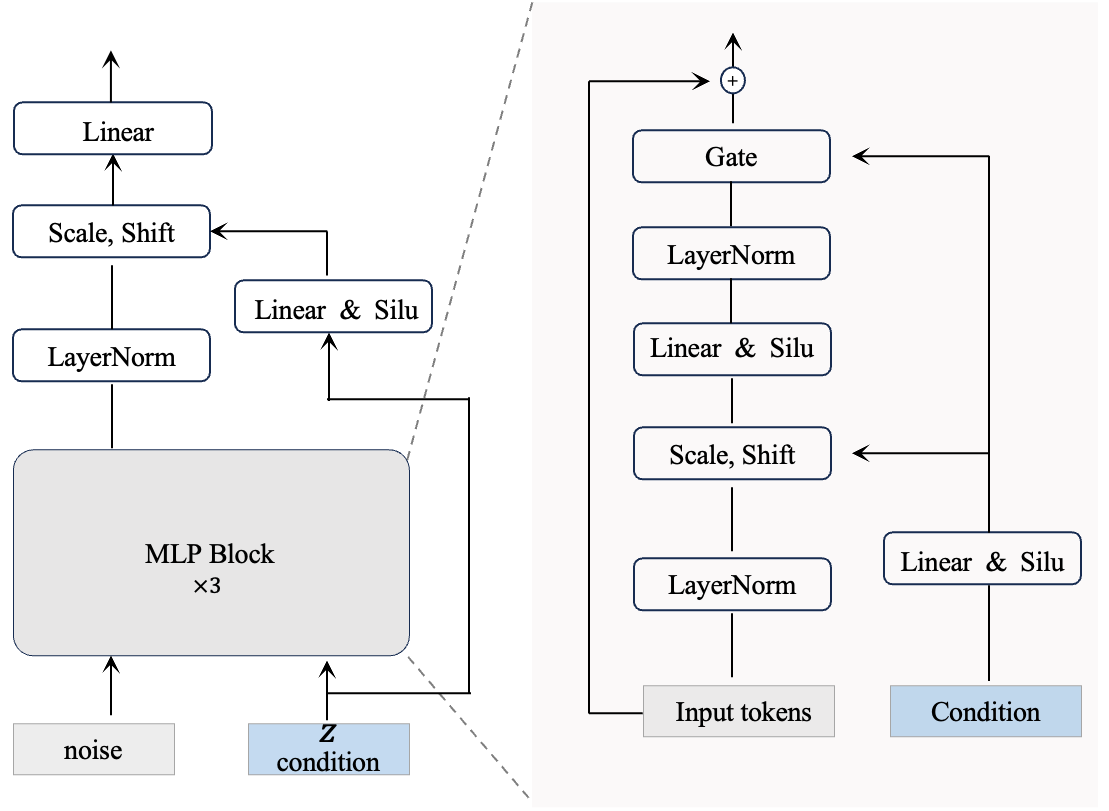}
   \vspace{-0.5cm}
   \caption{The structure of the DiffusionMLP.}
   \label{fig:diff}
\end{figure}

\vspace{-0.4cm}
\paragraph{Progressive Motion Prediction} In this PMP module, we first generate an outline feature for the current motion based on audio information.
We set the latest audios from three frames as conditions to provide more complete word pronunciation, and use the cross-attention to predict the coarse outline feature. Then, we utilize the contextual interaction summary (cis-token) and the state feature to perform fine-grained prediction via the condition block which includes cross-attention and feedforward function. We also combine $A_T^a$ to enhance the audio part. Then we refer to the 5-frame motions at the temporal layer to further enforce inter-frame continuity.
We then use the output of the temporal layer as a condition for denoising in the DiffusionMLP to sample motions. The DiffusionMLP is a lightweight network composed of 3 MLP blocks, with conditional injection also performed via adaLN\cite{dit}. The specific structure of DiffusionMLP is shown in \cref{fig:diff}.

\begin{figure*}[t]
   \includegraphics[width=\linewidth]{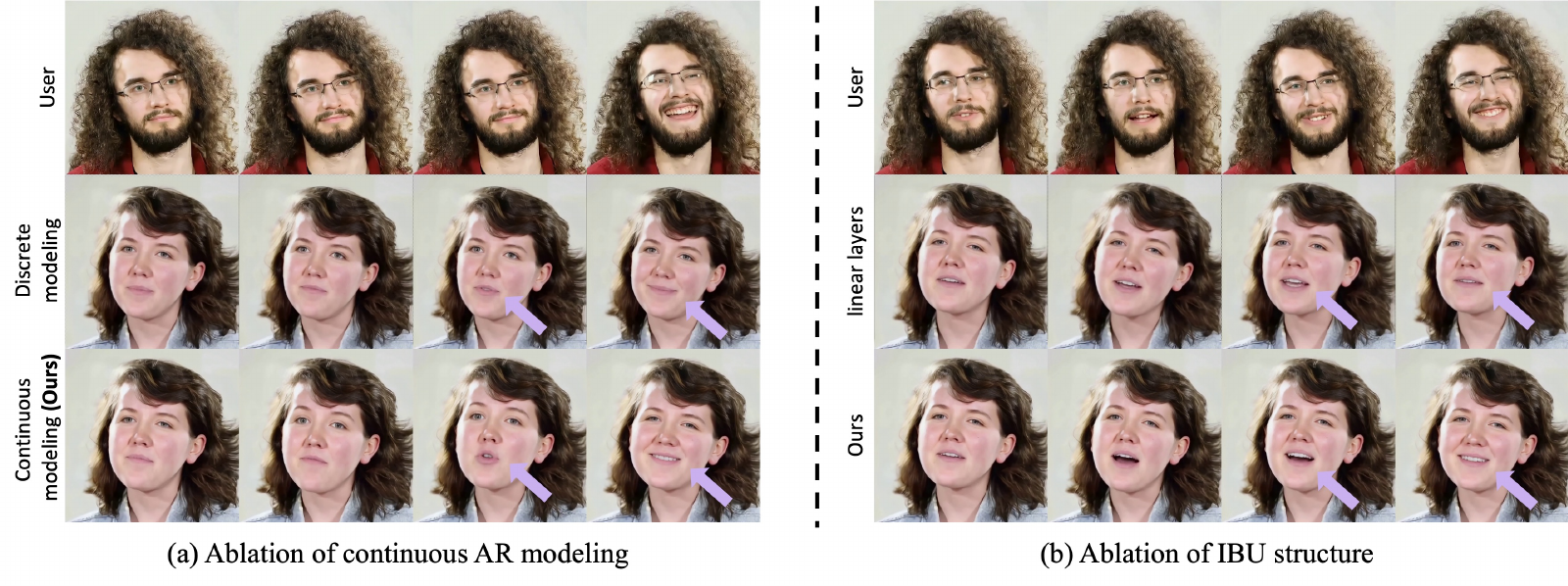}
   \caption{Ablation Study.}
   \label{fig:supp_ablation}
\end{figure*}

\begin{table*}[t]
\footnotesize
\centering

\begin{tabular}{c cc cc cc cc}
\toprule

    Methods & RPCC $\downarrow$ & CSIM $\uparrow$ & SyncScore $\uparrow$ & PSNR $\uparrow$ & SSIM $\uparrow$ & FID $\downarrow$ & SID$\uparrow$ & Var$\uparrow$  \\ \hline


    w/o continuous AR modeling   & 0.129 & 0.887 & 7.036 & 27.62 & 0.813 & 23.78 & 2.261 & 2.154  \\
    w/o bidirectional-integrated learning    & 0.173 & 0.841 & 6.813 & 24.17 & 0.749 & 25.36 & 2.138 & 2.016 \\

    Ours & \textbf{0.125} & \textbf{0.901} & \textbf{7.218} & \textbf{29.67} & \textbf{0.827} & \textbf{21.64} & \textbf{2.428} & \textbf{2.397}  \\ 
 
\bottomrule
\end{tabular}
\caption{Ablation study for continuous AR modeling as well as the bidirectional-integrated learning. \textbf{Bold} represents the best.}
\label{table:supp_ablation}
\end{table*}
\begin{table}[t]
\setlength{\tabcolsep}{4pt}
\footnotesize
\centering
\begin{tabular}{c c c c c}
\toprule
    \multirow{2}{*}{Methods}  & Overall  & User-agent  & Motion & Lip \\ 
    & Naturalness & Coordination & Diversity & Synchronization \\
    \hline
    DIM \cite{dim} &  2.48  &  2.04  &  2.12  &  2.57 \\
    Ours &  \textbf{4.43}  &  \textbf{4.18}  &  \textbf{4.52}  &  \textbf{4.36} \\
  
\bottomrule
\end{tabular}
\caption{User study. The best results are highlighted in bold.}
\label{table:user_study}
\end{table}

\subsection{Inference Details}
The dimensions of the input audio and motion are 768 and 262. We first encode them into 512 dimensions and put them into the historical input.
In the initial stage, we repeat the motion vector of the agent's reference image and the audio corresponding to the first frame to initialize each cache. The embedding dimension in Bidirectional-learning, Integrated-learning, Contextual-understanding and State-prediction is 512. The dim of feedforward function is 2048. The condition dim in DiffusionMLP is projected into 262.

\begin{figure*}[t]
   \includegraphics[width=\linewidth]{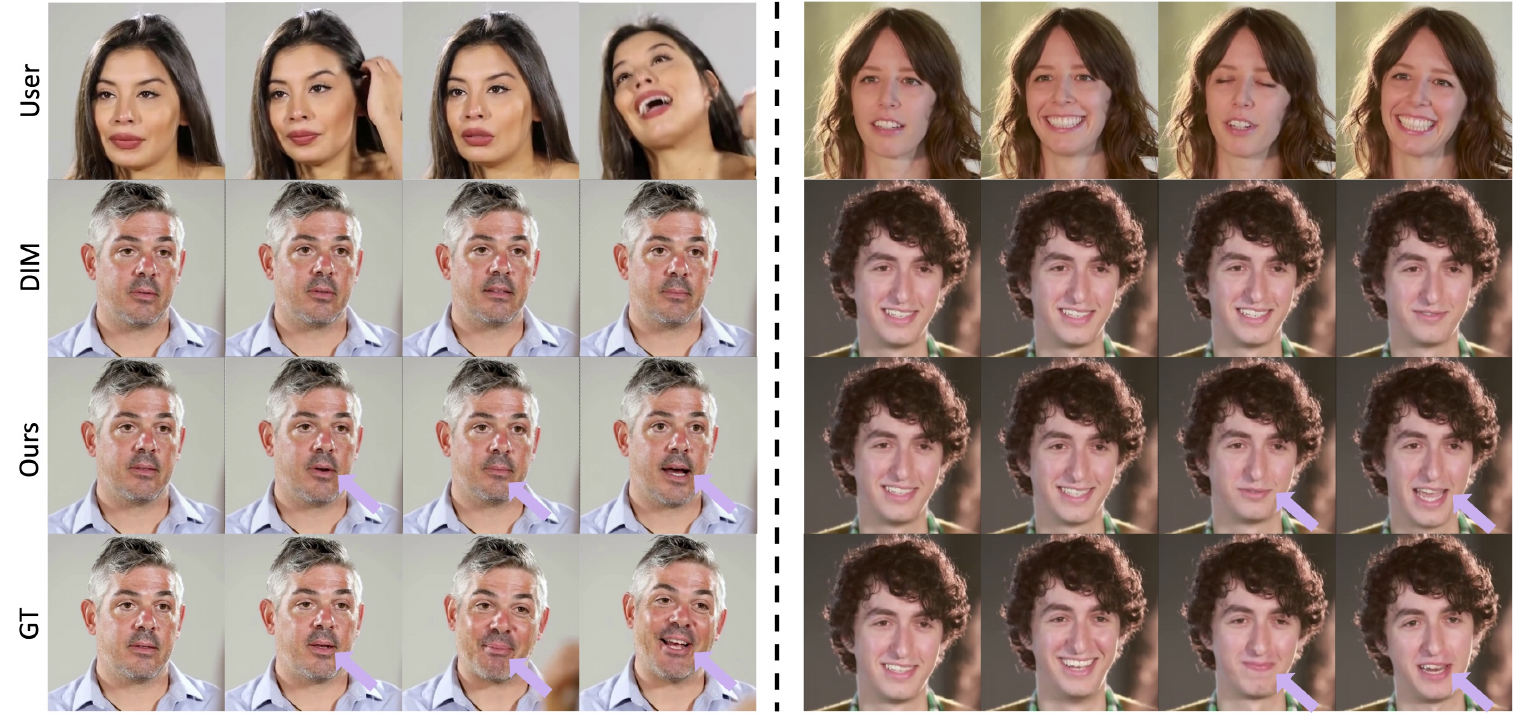}
   \caption{Qualitative comparisons with DIM\cite{dim} on RealTalk\cite{realtalk} dataset.}
   \label{fig:supp_inter}
\end{figure*}

\section{More Experiments}
\subsection{More Ablation}
\paragraph{Continuous AR Modeling}
To validate the effectiveness of continuous autoregressive (AR) modeling, we utilize discrete indices in an N-sized codebook to represent the motions, and employ the Softmax function as the sampler in PMP. The ablation results are shown in Table \ref{table:supp_ablation}, and the visual ablation is shown in Figure \ref{fig:supp_ablation} (a). As shown in Table \ref{table:supp_ablation}, metrics related to video realism (\eg FID, PSNR and SSIM) and motion diversity (\eg SID and Var) are greatly worsened when using discrete AR modeling. In Figure \ref{fig:supp_ablation} (a), we can also see that compared to continuous AR modeling, the discrete AR modeling may reduce facial expressiveness significantly, which demonstrates that the discrete modeling is not sufficient for the representation of rich facial expressions, and thus fails to predict realistic and diverse motions.

\paragraph{Bidirectional-integrated Learning}
We also ablate the network structure of our proposed IBU module and replace the bidirectional and integrated learning module with simple linear layers. As shown in Table \ref{table:supp_ablation}, all metrics have deteriorated significantly without bidirectional-integrated learning, which validates that our designed bidirectional and integrated learning module is effective in interactive behavior understanding and modeling. In Figure \ref{fig:supp_ablation} (b), we can see that our method can significantly improve the accuracy of facial motion prediction.

\subsection{User Study}
We asked 25 people to rate 20 different videos (on a scale of 1-5, the higher the better) generated by DIM\cite{dim} and our method across four dimensions: overall naturalness, user-agent coordination, motion diversity, and lip synchronization. To be specific, the 20 test videos are randomly chosen from RealTalk\cite{realtalk}. As shown in Table \ref{table:user_study}, our method outperforms DIM\cite{dim} in all aspects.

\begin{figure*}[t]
   \includegraphics[width=\linewidth]{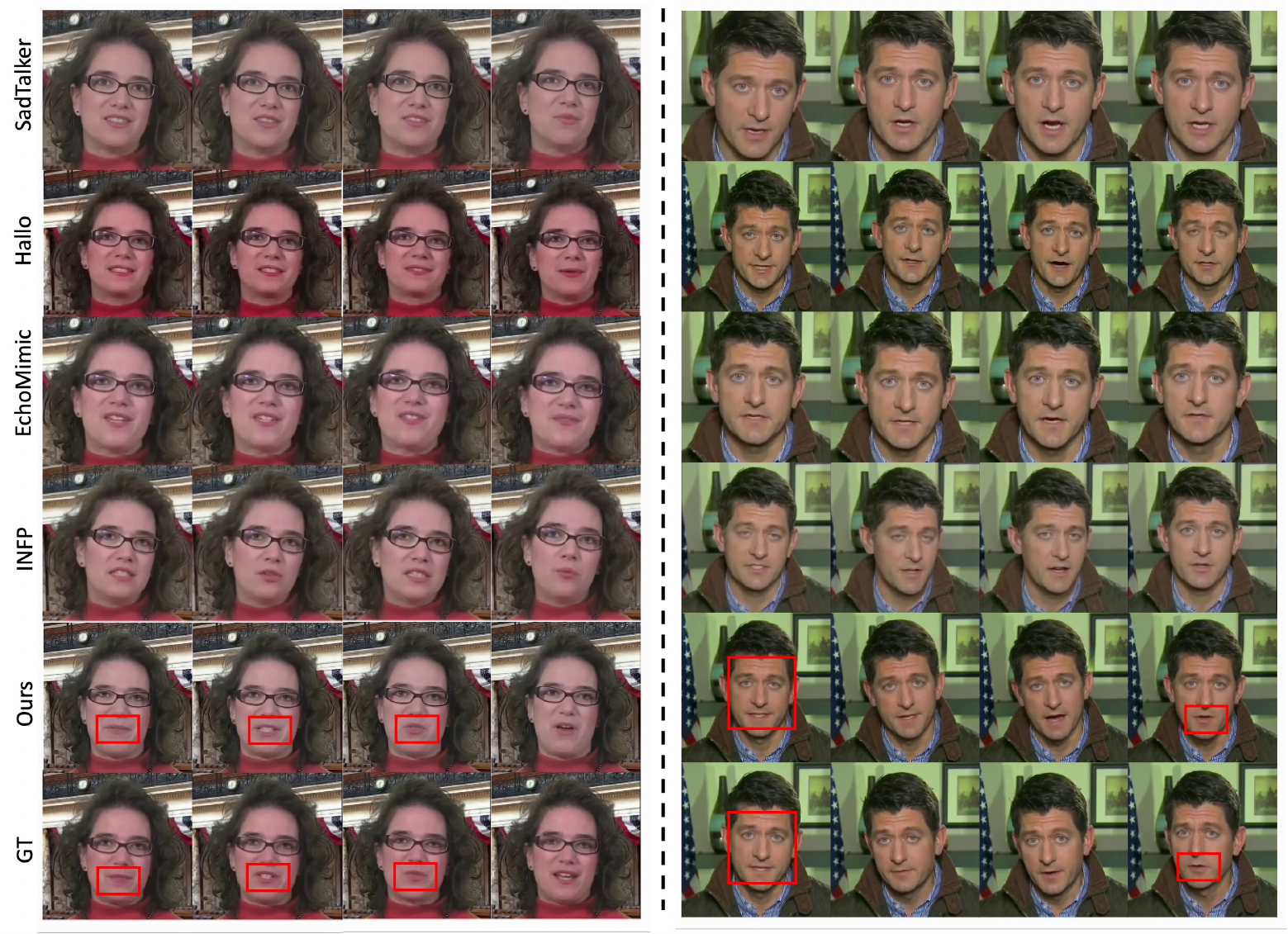}
   \caption{Qualitative comparisons with state-of-the-art talking head generation methods on HDTF\cite{hdtf} dataset.}
   \label{fig:supp_talk}
\end{figure*}

\begin{figure*}[t]
   \includegraphics[width=\linewidth]{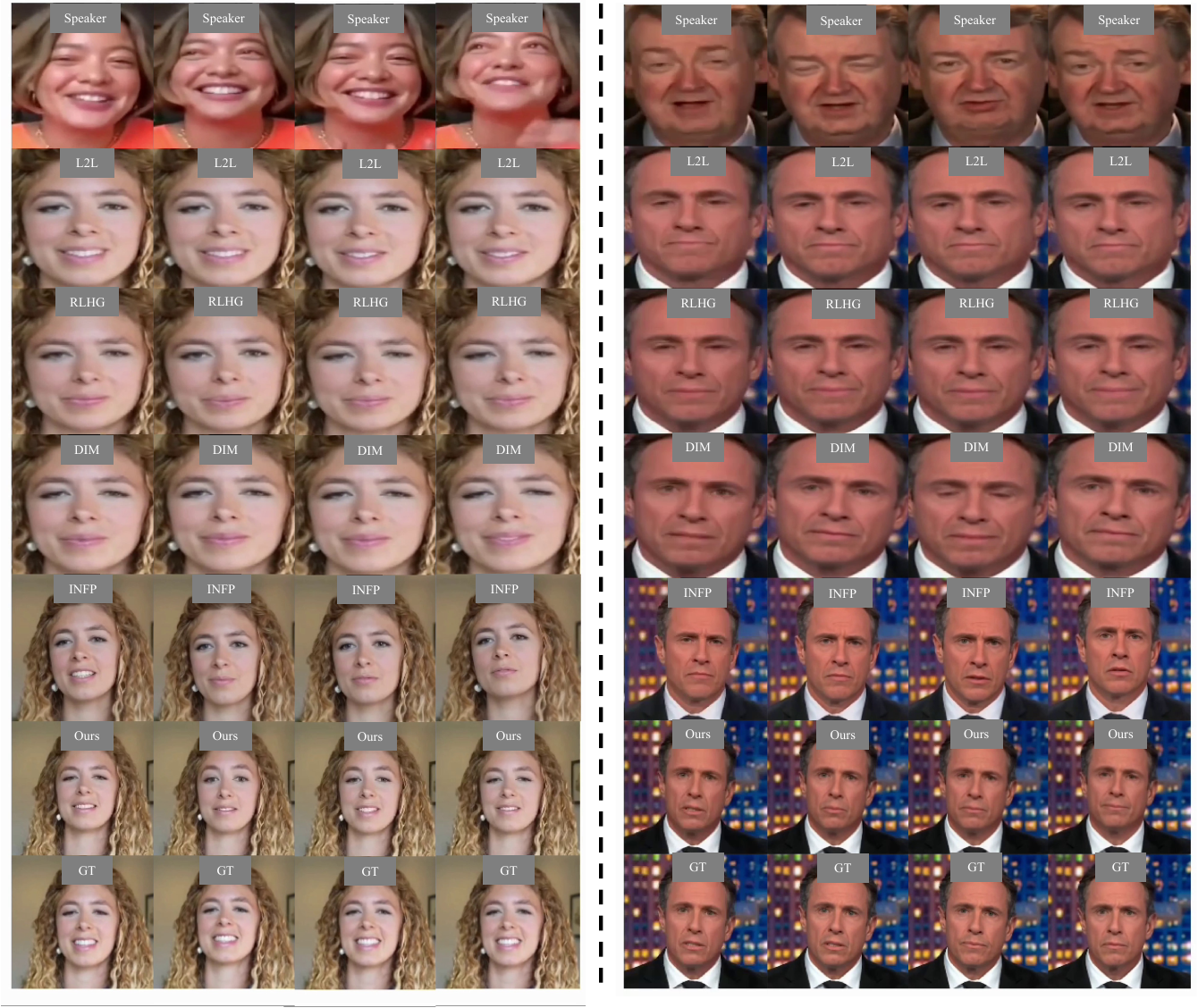}
   \caption{Qualitative comparisons with state-of-the-art listenning head generation methods on ViCo\cite{vico} dataset.}
   \label{fig:supp_listen}
\end{figure*}

\section{Supplementary Visual Results}
\subsection{Interactive Head Generation}
We present visual comparisons of our method with DIM\cite{dim} on RealTalk\cite{realtalk} dataset in Figure \ref{fig:supp_inter}. It can be seen that the agent videos generated by our method are closer to GT and have a significant improvement compared to DIM\cite{dim}.

\subsection{Talking Head Generation}
In Figure \ref{fig:supp_talk}, we compare our results with SadTalker\cite{zhang2023sadtalker}, Hallo\cite{xu2024hallo} and EchoMimic\cite{chen2024echomimic} on HDTF\cite{hdtf} based on the same reference image and audio. It can be seen that compared to other methods, our method exhibits the closest lip movements and head motions to the ground truth, indicating that the motions produced
by our method are the most natural and photorealistic.

\subsection{Listening Head Generation}
We compare our method with the SOTA listening head generation methods (e.g., RLHG\cite{vico}, L2L\cite{l2l}, DIM\cite{dim} and INFP\cite{infp}) on ViCo\cite{vico} based on the same speaker and reference image. As shown in Figure \ref{fig:supp_listen}, the results generated by our method are the most similar to the ground-truth videos, demonstrating great superiority in the realism of facial motions.

\section{Limitations and Social Impact}
Although our method can achieve real-time interactive motion generation, its application scope is limited to the head and cannot cover the body generation, which deserves further research in the future.
In practical applications, our generation method has many positive effects, such as the character animation in movies, creating virtual hosts for advertising, and developing interactive teaching tools to provide immersive experience.
However, it may be abused in some scenarios (\eg creating false content for bullying others, creating deceptive videos for spreading misinformation.).
In order to prevent the technology from being abused, we can ensure that the technology serves a positive purpose by forcibly adding watermarks to the generated content and managing the way the code is obtained.

\end{document}